\documentclass{article}

\usepackage{PRIMEarxiv}

\usepackage[utf8]{inputenc} 
\usepackage[T1]{fontenc}    
\usepackage{hyperref}       
\usepackage{url}            
\usepackage{booktabs}       
\usepackage{amsfonts}       
\usepackage{nicefrac}       
\usepackage{microtype}      
\usepackage{lipsum}
\usepackage{fancyhdr}       
\usepackage{graphicx}       
\graphicspath{{media/}}     

\usepackage{multirow}
\usepackage[table]{xcolor}
\usepackage{float}
\usepackage{tcolorbox}
\tcbuselibrary{breakable}
\usepackage{listings}
\usepackage{longtable}
\usepackage{array}

\title{UltraSAM3: A Concept-Driven Foundation Model for Universal Ultrasound Image Segmentation
}

\author{
  Bo Xu, Quanhao Zhu, Rui Lin, Boling Zhu, Chenyuan Wang, Hongfei Lin \\
  School of Software \\
  Dalian University of Technology \\
  \texttt{boxu@dlut.edu.cn} \\
  \texttt{zhuqh19@gmail.com} \\
  \texttt{lr983088162@mail.dlut.edu.cn} \\
  \texttt{SinsapZ@outlook.com} \\
  \texttt{chenyuanwang@mail.dlut.edu.cn} \\
  \texttt{hflin@dlut.edu.cn}
  \AND
  Feng Xia \\
  School of Computing Technologies \\
  RMIT University \\
  \texttt{f.xia@ieee.org}
  \And
  Chenhua Ji \\
  Dalian University of Technology Affiliated Center Hospital \\
  \texttt{chenhuaji1974@126.com}
}

\begin{document}
\maketitle

\begin{abstract}
Ultrasound imaging has become increasingly widespread in clinical practice due to its portability, low cost and real-time capability, making ultrasound image segmentation important. However, ultrasound images differ substantially from CT, MRI, and other medical imaging modalities, as they are often affected by speckle noise, low contrast, acoustic shadows and ambiguous boundaries. Existing ultrasound segmentation methods are still mainly limited to task-specific models or visual-prompt-based foundation models, which are either tailored to particular tasks or require expert-provided visual prompts, making them inconvenient for flexible clinical use. To address these challenges, we propose UltraSAM3, a concept-driven foundation model for universal ultrasound image segmentation. Unlike conventional models, UltraSAM3 enables text-based target specification by adapting SAM3 to ultrasound-specific image--mask--concept triplets. The model is trained on a large-scale ultrasound segmentation corpus covering 37 public datasets and 13 anatomical categories, allowing it to align ultrasound visual patterns with clinically meaningful concepts across diverse organs and lesions. To further improve usability under realistic clinical interaction, we propose an instruction-guided agent that parses complex natural language queries into concise ultrasound concept prompts for UltraSAM3. Extensive experiments demonstrate that UltraSAM3 consistently outperforms representative concept- and text-driven biomedical segmentation models on multi-organ ultrasound benchmarks, external datasets, and visual-prompt-enhanced settings. Moreover, the agent improves segmentation robustness for complex user instructions. These results indicate that ultrasound-specific concept adaptation is effective for building generalizable and interactive ultrasound segmentation foundation models. Our code is available at \url{https://github.com/zhuqh19/UltraSAM3}.
\end{abstract}


\section{Introduction}

Ultrasound image segmentation aims to delineate clinically relevant anatomical structures and lesions from ultrasound images at the pixel level. As a non-invasive, radiation-free, real-time, and cost-effective imaging modality, ultrasound is widely used in routine clinical practice, including fetal assessment, cardiac examination, breast lesion screening, thyroid nodule analysis, liver and kidney evaluation, vascular imaging, nerve localization, and musculoskeletal diagnosis. Accurate segmentation of ultrasound targets enables objective anatomical measurement, lesion quantification, treatment planning, image-guided intervention, and longitudinal follow-up. However, compared with other medical imaging modalities, ultrasound images exhibit distinctive visual characteristics, such as speckle noise, acoustic shadowing, low tissue contrast, ambiguous boundaries, view-dependent appearance, and strong operator dependence. These factors make ultrasound segmentation particularly challenging and limit the generalization ability of models trained on individual datasets or narrowly defined tasks.

Existing ultrasound segmentation methods are still largely dominated by task-specific supervised learning. Classical convolutional architectures such as U-Net \cite{ronneberger2015u} and automated pipelines such as nnU-Net \cite{isensee2021nnu} have achieved strong performance on many medical segmentation benchmarks. Transformer-based models further improve global context modeling, as exemplified by TransUNet \cite{chen2021transunet} and UNETR \cite{hatamizadeh2022unetr}, while recent Mamba-based models explore efficient long-range dependency modeling for medical segmentation \cite{ma2024u,ruan2024vm,liu2024swin,dang2024log}. Despite these advances, most existing methods are optimized for fixed label spaces and specific imaging distributions. In ultrasound imaging, this limitation is especially severe because different organs, scanners, acquisition protocols, and clinical tasks often lead to substantial appearance shifts. As a result, separate models are usually required for fetal head segmentation, thyroid nodule segmentation, breast lesion segmentation, cardiac chamber segmentation, vascular structure segmentation, and other ultrasound tasks. Such a task-specific paradigm is difficult to scale to diverse clinical scenarios and cannot flexibly respond to natural language instructions or open-set medical concepts.

\begin{figure*}[t]
    \centering
    \includegraphics[width=\textwidth]{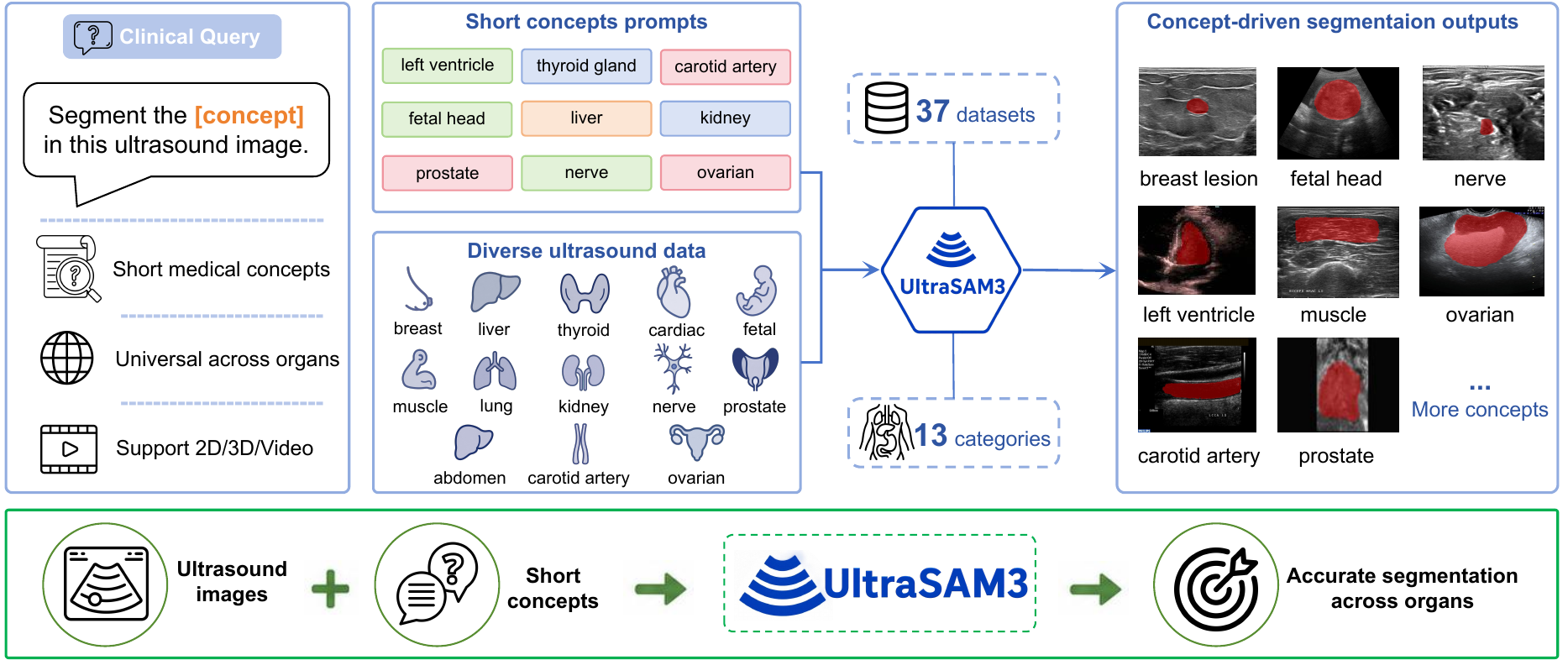}
    \caption{Overview of UltraSAM3 for concept-driven universal ultrasound segmentation. Given ultrasound images and short medical concepts, UltraSAM3 produces accurate segmentation outputs across diverse organs and datasets.}
    \label{fig:introduction}
\end{figure*}

Promptable segmentation foundation models provide a promising alternative by enabling segmentation through user-specified prompts rather than fixed output categories. The Segment Anything Model (SAM) \cite{kirillov2023segment} introduces a general promptable segmentation interface based on points, boxes, and masks, and SAM2 \cite{ravi2024sam} extends this paradigm to video and sequential image settings. Medical adaptations such as MedSAM \cite{ma2024segment} and MedSAM2 \cite{ma2025medsam2} further demonstrate the potential of promptable segmentation in medical images. For ultrasound imaging, recent efforts such as UltraSAM \cite{meyer2025ultrasam} and SAMUSA \cite{podvin2025samusa} indicate that ultrasound-specific adaptation can improve annotation efficiency and segmentation robustness. However, most of these models still rely on visual or geometric prompts, such as bounding boxes, points, or masks. In practical clinical usage, such prompts are not always available: requiring a user to provide a precise box or point assumes that the target has already been localized. This weakens the autonomy of the segmentation system and limits its ability to handle more natural clinical requests, such as ``segment the fetal head'', ``find the thyroid nodule'', or ``delineate the suspicious breast lesion''.

Recent concept- and text-prompted segmentation models aim to reduce this dependence on geometric prompts. SAM3 \cite{carion2025sam} advances the segmentation paradigm from visual prompting to concept-based prompting, enabling segmentation according to phrase-level semantic concepts. In the biomedical domain, BiomedParse \cite{zhao2024biomedparse}, SAT \cite{zhao2025large}, UniBiomed \cite{wu2025unibiomed}, MedSAM3 \cite{liu2025medsam3}, and Medical SAM3 \cite{jiang2026medical} explore text-driven or concept-driven medical image understanding and segmentation. Nevertheless, most existing concept-prompted medical models are designed for general biomedical or multi-modal medical imaging scenarios. Ultrasound remains under-explored as a dedicated domain. Unlike CT, MRI, pathology, or endoscopy, ultrasound images are strongly affected by acoustic artifacts, unstable scanning planes, low signal-to-noise ratio, and large intra-class appearance variations. These characteristics can severely weaken semantic grounding: a model may understand the textual concept but fail to localize the corresponding structure in noisy and ambiguous ultrasound images. Therefore, a dedicated ultrasound foundation model is needed to align ultrasound-specific visual patterns with medical concepts across organs, lesions, and acquisition conditions.

To address these challenges, we propose \textbf{UltraSAM3}, a concept-driven foundation model for universal ultrasound image segmentation. As shown in Fig.~\ref{fig:introduction}, UltraSAM3 focuses on ultrasound-centric representation learning and semantic grounding. We adapt SAM3 to a large-scale ultrasound segmentation corpus consisting of 37 public ultrasound datasets covering 13 organ and anatomical categories, including lung, muscle, abdomen, ovary, heart, prostate, thyroid, liver, fetus, nerve, breast, carotid artery, and kidney. Each training sample is formulated as an image--mask--concept triplet, where the textual concept is derived from the corresponding organ, lesion, or anatomical label. Through this formulation, UltraSAM3 learns to associate ultrasound-specific visual appearances with standardized medical concepts, enabling segmentation driven by text prompts without relying on privileged spatial cues such as ground-truth bounding boxes. However, complex user instructions may degrade the segmentation performance of UltraSAM3. To address this problem, we further build a simple instruction-guided agent on top of UltraSAM3, which converts complex user queries into explicit ultrasound concept prompts and produces segmentation-grounded responses, providing a more accessible interface for concept-driven ultrasound segmentation.

In summary, our contributions are threefold:
\begin{itemize}
    \item We propose \textbf{UltraSAM3}, to the best of our knowledge, the first ultrasound-specific concept-driven segmentation foundation model that adapts to challenging ultrasound imaging scenarios and supports text-based target specification without visual prompts.
    
    \item We develop an instruction-guided ultrasound segmentation agent that converts complex natural language instructions into short ultrasound concept prompts and uses the predicted masks to support segmentation-grounded question answering.
    
    \item We conduct extensive experiments across diverse ultrasound datasets and anatomical categories, demonstrating that UltraSAM3 achieves state-of-the-art performance on multiple ultrasound segmentation benchmarks and exhibits strong generalization under text-only prompting.
\end{itemize}

\section{Related Work}

\subsection{Task-specific Medical Image Segmentation}
Early medical image segmentation methods mainly followed a task-specific paradigm, where separate models were designed and trained for particular organs, lesions, or imaging modalities. U-Net \cite{ronneberger2015u} is one of the most representative architectures in this line of research, using an encoder-decoder structure with skip connections to effectively integrate semantic representations and spatial details. nnU-Net \cite{isensee2021nnu} further standardized task-specific segmentation pipelines by automatically configuring preprocessing, network architecture, training strategies, and post-processing, achieving strong robustness across diverse medical segmentation tasks. Later studies introduced Transformers to enhance global context modeling. For example, TransUNet \cite{chen2021transunet} combines a Transformer encoder with a U-Net-like decoder, while UNETR \cite{hatamizadeh2022unetr} formulates 3D medical image segmentation as a Transformer-based sequence learning problem. More recently, Mamba and state space models have been explored for medical segmentation. U-Mamba \cite{ma2024u}, VM-UNet \cite{ruan2024vm}, Swin-UMamba \cite{liu2024swin}, and LoG-VMamba \cite{dang2024log} improve segmentation performance by enhancing long-range dependency modeling, efficient feature extraction, and local-global representation learning. Although these models achieve strong performance on specific datasets and tasks, they usually rely on fixed category annotations and task-specific training, making them difficult to generalize to unseen organs, lesions, or fine-grained semantic categories. Moreover, they cannot flexibly perform segmentation according to natural language instructions or open-set concept prompts. Therefore, in ultrasound medical imaging, task-specific models still suffer from limited generalizability and insufficient interactive capability.

\subsection{Visual-prompted Medical Image Segmentation}
With the introduction of the Segment Anything Model (SAM) \cite{kirillov2023segment}, medical image segmentation has gradually shifted from conventional fixed-category prediction\cite{li2024kd,xu2025echogpt} toward a promptable segmentation paradigm. Trained on large-scale data, SAM can generate segmentation masks conditioned on visual prompts such as points, boxes, and masks, substantially improving interactivity and generalization. SAM2 \cite{ravi2024sam} further extends this paradigm from images to videos by introducing a memory mechanism for object tracking and continuous segmentation across frames. Inspired by these advances, a series of visual-prompt-based SAM adaptations have been developed for medical imaging. MedSAM \cite{ma2024segment} fine-tunes SAM on large-scale medical image segmentation datasets, making it more suitable for multi-modal medical images; however, it primarily relies on box prompts and therefore still requires accurate object localization from human users or external models. MedSAM2 \cite{ma2025medsam2} extends SAM2 to 3D medical images and medical videos, improving its applicability to volumetric and temporal medical data. Dong et al. \cite{dong2026segment} systematically evaluate SAM2 on 2D and 3D medical images, showing that general-purpose visual segmentation models still require domain adaptation in medical scenarios. For ultrasound imaging, UltraSAM \cite{meyer2025ultrasam} and SAMUSA \cite{podvin2025samusa} explore foundation models trained on large-scale ultrasound segmentation datasets and SAM2-based ultrasound annotation, respectively, demonstrating the potential of promptable foundation models in ultrasound applications. Nevertheless, ultrasound images often suffer from speckle noise, low contrast, ambiguous boundaries, and large anatomical variations. Geometric prompts such as points, boxes, and masks are insufficient to fully express the semantic meaning of medical targets. In addition, these methods typically require manual interaction or additional detectors to generate prompts, limiting their applicability to complex natural language instructions and open-concept segmentation.

\subsection{Text- and Concept-prompted Medical Image Segmentation}
To reduce the dependence on manual geometric prompts, recent studies have begun to explore text-prompted or concept-prompted medical image segmentation. SAM3 \cite{carion2025sam} advances the segmentation paradigm from promptable visual segmentation to promptable concept segmentation, enabling models to segment target instances according to phrase-level concepts, visual exemplars, or their combination. This provides a new foundation for open-vocabulary and natural-language-driven segmentation. In the biomedical domain, BiomedParse \cite{zhao2024biomedparse} attempts to build a unified biomedical image parsing model that integrates recognition, detection, and segmentation into a single framework, supporting multi-class and multi-task biomedical image understanding. SAT \cite{zhao2025large} further investigates large-vocabulary medical image segmentation with text prompts, allowing models to segment targets based on medical terminology. UniBiomed \cite{wu2025unibiomed} focuses on grounded biomedical image interpretation and uses text prompts to perform multiple biomedical image understanding tasks, including referring segmentation. Following the emergence of SAM3, MedSAM3 \cite{liu2025medsam3} and Medical SAM3 \cite{jiang2026medical} introduce concept-prompted segmentation into medical imaging and explore universal prompt-driven segmentation based on medical concepts. Although these methods substantially improve semantic interaction in medical segmentation, most of them are designed for general medical imaging or multi-modal scenarios, and their adaptation to ultrasound remains insufficient. Ultrasound images are characterized by noise, artifacts, low contrast, and uncertain boundaries, which can weaken the localization ability of general text-prompted models. Moreover, many ultrasound tasks involve multi-organ recognition and fine-grained medical concepts such as benign, malignant, and normal categories, posing greater challenges to existing general-purpose text-prompted segmentation models. Therefore, it is necessary to develop a SAM3-based model specifically adapted to ultrasound imaging and to incorporate a natural language instruction parsing mechanism that converts complex user queries into stable and explicit medical concept prompts.

\begin{figure*}[t]
    \centering
    \includegraphics[width=0.96\textwidth]{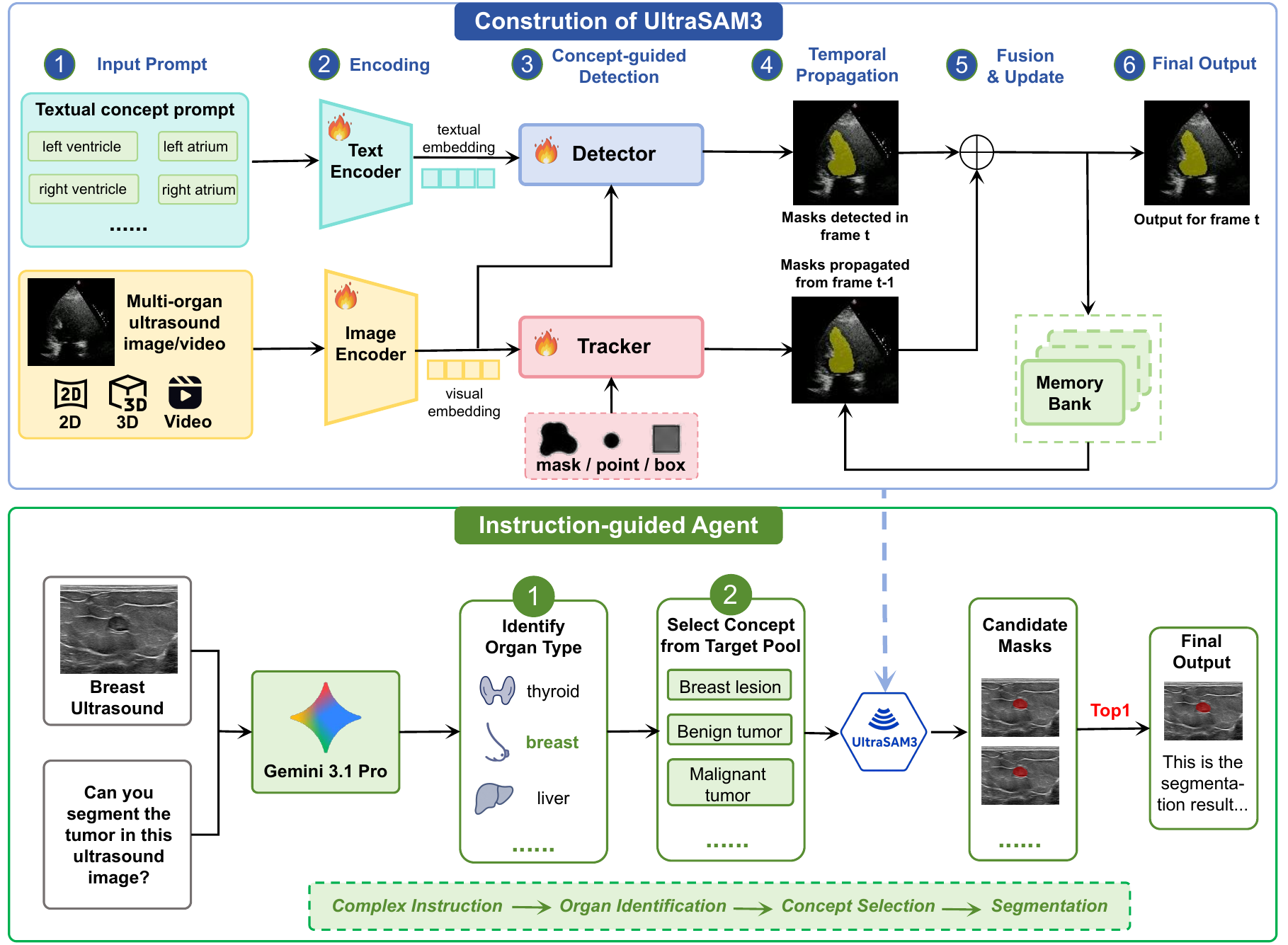}
    \caption{Overall framework of UltraSAM3 and the instruction-guided agent. The upper part illustrates the construction of UltraSAM3 through full-parameter fine-tuning of SAM3 with ultrasound images/videos and textual medical concepts. The lower part shows the agent-assisted inference pipeline, where Gemini 3.1 Pro first identifies the organ type from a complex instruction and ultrasound image, selects the target concept from the corresponding organ-specific concept pool, and then sends the parsed prompt to UltraSAM3 for candidate mask generation and final segmentation.}
    \label{fig:framework}
\end{figure*}

\section{Method}

\subsection{Problem Definition}
Given an ultrasound image or frame $I \in \mathbb{R}^{H \times W \times 3}$ and a textual concept prompt $q$, the goal of concept-driven ultrasound segmentation is to predict a binary mask $M \in \{0,1\}^{H \times W}$ that delineates the target anatomical structure or lesion described by $q$. In the training set, each sample is formulated as an image--mask--concept triplet
\begin{equation}
    \mathcal{D} = \{(I_i, M_i, q_i)\}_{i=1}^{N},
\end{equation}
where $I_i$ denotes the ultrasound image, $M_i$ is the ground-truth segmentation mask, and $q_i$ is the corresponding medical concept, such as an organ, anatomical structure, or lesion category. For datasets containing multiple semantic categories, each annotated object is converted into an individual concept-conditioned segmentation sample.

The model learns a concept-conditioned segmentation function
\begin{equation}
    \hat{M}_i = f_{\theta}(I_i, q_i),
\end{equation}
where $f_{\theta}$ denotes UltraSAM3 with trainable parameters $\theta$, and $\hat{M}_i$ is the predicted mask. During inference, the model receives a user-specified concept prompt $q$ and produces one or more candidate masks with confidence scores:
\begin{equation}
    \{(\hat{M}^{k}, s^{k})\}_{k=1}^{K} = f_{\theta}(I, q),
\end{equation}
where $K$ is the number of predicted mask candidates and $s^{k}$ denotes the confidence score of the $k$-th mask. The final segmentation result is selected according to the highest confidence score:
\begin{equation}
    \hat{M} = \hat{M}^{k^{*}}, \quad
    k^{*} = \arg\max_{k} s^{k}.
\end{equation}

\subsection{Framework Overview}
As shown in Fig.~\ref{fig:framework}, the overall framework of UltraSAM3 consists of two stages: concept-driven segmentation model adaptation and instruction-guided inference. In the first stage, SAM3 is adapted to ultrasound images using paired ultrasound images, segmentation masks, and textual medical concepts. The textual prompt $q$ is encoded by a text encoder, while the ultrasound image $I$ is processed by an image encoder. The encoded visual and textual representations are then fed into the segmentation modules of SAM3 to generate concept-conditioned masks. This design enables the model to associate ultrasound-specific visual patterns with clinically meaningful concepts.

In the second stage, UltraSAM3 can be used either directly with a concise concept prompt or together with an instruction-guided agent. For simple clinical concepts, the user prompt is directly used as the input query. For complex natural-language instructions, the agent first parses the user query, identifies the intended target category within the current organ domain, and rewrites the instruction into a compact SAM3-compatible concept prompt. The parsed prompt is then passed to the fine-tuned UltraSAM3 to produce the final segmentation mask. Therefore, the framework decomposes complex user interaction into semantic query parsing and concept-driven mask prediction:
\begin{equation}
    \hat{M} =
    f_{\theta}\big(I, g_{\phi}(I, u, \mathcal{C}_{o})\big),
\end{equation}
where $u$ denotes a complex user instruction, $\mathcal{C}_{o}$ is the organ-level candidate concept set, and $g_{\phi}$ is the instruction-guided agent that outputs a simplified concept prompt.

\begin{table*}[t]
\centering
\caption{Performance comparison on representative datasets from 13 ultrasound categories. The best result is shown in bold, the second-best result is underlined, and $\Delta$ denotes the absolute improvement of UltraSAM3 over the strongest baseline.}
\label{tab:representative_dataset_results}
\setlength{\tabcolsep}{4pt}
\renewcommand{\arraystretch}{1.08}
\resizebox{\textwidth}{!}{%
\begin{tabular}{ll l ccccc c}
\toprule
\rowcolor{gray!15}
\textbf{Category} & \textbf{Dataset} & \textbf{Metric} & \textbf{UniBiomed} & \textbf{BiomedParse} & \textbf{SAM3} & \textbf{Medical SAM3} & \textbf{UltraSAM3} & \textbf{$\Delta$} \\
\midrule

\multirow{2}{*}{Abdomen} & \multirow{2}{*}{AbdomenUS\cite{orlando_abdomenus_ussimandsegm}} 
& IoU  & 0.0646 & 0.0802 & 0.1224 & \underline{0.1916} & \textbf{0.4163} & \textcolor{green!50!black}{+0.2247} \\
& & Dice & 0.0989 & 0.1093 & 0.1779 & \underline{0.2672} & \textbf{0.5118} & \textcolor{green!50!black}{+0.2446} \\
\addlinespace[2pt]

\multirow{2}{*}{Breast} & \multirow{2}{*}{BUS\_BRA\cite{gomez2024bus}} 
& IoU  & 0.4981 & 0.6989 & 0.4522 & \underline{0.7293} & \textbf{0.8156} & \textcolor{green!50!black}{+0.0863} \\
& & Dice & 0.6084 & 0.7975 & 0.5054 & \underline{0.8091} & \textbf{0.8838} & \textcolor{green!50!black}{+0.0747} \\
\addlinespace[2pt]

\multirow{2}{*}{Cardiac} & \multirow{2}{*}{EchoNet\_Dynamic\cite{ouyang2020video}} 
& IoU  & 0.4291 & 0.2579 & 0.1119 & \underline{0.5089} & \textbf{0.8463} & \textcolor{green!50!black}{+0.3374} \\
& & Dice & 0.5690 & 0.3376 & 0.1514 & \underline{0.6121} & \textbf{0.9145} & \textcolor{green!50!black}{+0.3024} \\
\addlinespace[2pt]

\multirow{2}{*}{Carotid\_artery} & \multirow{2}{*}{CCA\cite{bi2023mi}} 
& IoU  & 0.0133 & \underline{0.4531} & 0.0291 & 0.4479 & \textbf{0.8245} & \textcolor{green!50!black}{+0.3714} \\
& & Dice & 0.0190 & 0.5280 & 0.0374 & \underline{0.5354} & \textbf{0.8998} & \textcolor{green!50!black}{+0.3644} \\
\addlinespace[2pt]

\multirow{2}{*}{Fetal} & \multirow{2}{*}{fh\_ps\cite{jieyun_2024_10829116}} 
& IoU  & 0.0025 & 0.0000 & 0.0455 & \underline{0.7624} & \textbf{0.7831} & \textcolor{green!50!black}{+0.0207} \\
& & Dice & 0.0042 & 0.0000 & 0.0773 & \underline{0.8637} & \textbf{0.8680} & \textcolor{green!50!black}{+0.0043} \\
\addlinespace[2pt]

\multirow{2}{*}{Kidney} & \multirow{2}{*}{Ultrasound\_Normal\_Kidney\cite{jeevaws2025ultrasoundnormalkidney}} 
& IoU  & 0.0380 & 0.2082 & \underline{0.4787} & 0.3267 & \textbf{0.5814} & \textcolor{green!50!black}{+0.1027} \\
& & Dice & 0.0720 & 0.3287 & \underline{0.6425} & 0.4738 & \textbf{0.7271} & \textcolor{green!50!black}{+0.0846} \\
\addlinespace[2pt]

\multirow{2}{*}{Liver} & \multirow{2}{*}{Annotated\_Ultrasound\_Liver\cite{xu_yiming_2022_7272660}} 
& IoU  & 0.1243 & 0.1941 & 0.0790 & \underline{0.2050} & \textbf{0.4454} & \textcolor{green!50!black}{+0.2404} \\
& & Dice & 0.1785 & 0.2455 & 0.0992 & \underline{0.2546} & \textbf{0.5099} & \textcolor{green!50!black}{+0.2553} \\
\addlinespace[2pt]

\multirow{2}{*}{Lung} & \multirow{2}{*}{LUSS\cite{mclaughlan2024lung}} 
& IoU  & 0.0219 & \underline{0.0595} & 0.0041 & 0.0386 & \textbf{0.1325} & \textcolor{green!50!black}{+0.0730} \\
& & Dice & 0.0332 & \underline{0.0914} & 0.0077 & 0.0713 & \textbf{0.1936} & \textcolor{green!50!black}{+0.1022} \\
\addlinespace[2pt]

\multirow{2}{*}{Muscle} & \multirow{2}{*}{STMUS\_NDA\cite{marzola2021deep}} 
& IoU  & 0.1465 & \underline{0.2264} & 0.0012 & 0.0871 & \textbf{0.8023} & \textcolor{green!50!black}{+0.5759} \\
& & Dice & 0.2142 & \underline{0.3019} & 0.0017 & 0.1371 & \textbf{0.8818} & \textcolor{green!50!black}{+0.5799} \\
\addlinespace[2pt]

\multirow{2}{*}{Nerve} & \multirow{2}{*}{UPBD\cite{ding2022mallesnet}} 
& IoU  & 0.0067 & 0.1190 & 0.1239 & \underline{0.1402} & \textbf{0.2031} & \textcolor{green!50!black}{+0.0629} \\
& & Dice & 0.0109 & 0.1442 & 0.2070 & \underline{0.2174} & \textbf{0.2923} & \textcolor{green!50!black}{+0.0749} \\
\addlinespace[2pt]

\multirow{2}{*}{Ovarian} & \multirow{2}{*}{OTU\_2d\cite{zhao2022mmotu}} 
& IoU  & 0.3199 & \underline{0.5289} & 0.1099 & 0.4843 & \textbf{0.7356} & \textcolor{green!50!black}{+0.2067} \\
& & Dice & 0.4307 & \underline{0.6157} & 0.1414 & 0.5717 & \textbf{0.8137} & \textcolor{green!50!black}{+0.1980} \\
\addlinespace[2pt]

\multirow{2}{*}{Prostate} & \multirow{2}{*}{MicroSeg\cite{shao2024micro}} 
& IoU  & \underline{0.4698} & 0.3303 & 0.0111 & 0.3051 & \textbf{0.8732} & \textcolor{green!50!black}{+0.4034} \\
& & Dice & \underline{0.6002} & 0.4218 & 0.0193 & 0.3965 & \textbf{0.9222} & \textcolor{green!50!black}{+0.3220} \\
\addlinespace[2pt]

\multirow{2}{*}{Thyroid} & \multirow{2}{*}{TG3K\cite{gong2023thyroid}} 
& IoU  & \underline{0.0273} & 0.0066 & 0.0131 & 0.0034 & \textbf{0.7857} & \textcolor{green!50!black}{+0.7584} \\
& & Dice & \underline{0.0390} & 0.0109 & 0.0236 & 0.0058 & \textbf{0.8687} & \textcolor{green!50!black}{+0.8297} \\
\addlinespace[2pt]

\midrule
\multirow{2}{*}{Average} & \multirow{2}{*}{--} 
& IoU  & 0.1663 & 0.2433 & 0.1217 & \underline{0.3254} & \textbf{0.6342} & \textcolor{green!50!black}{+0.3088} \\
& & Dice & 0.2214 & 0.3025 & 0.1609 & \underline{0.4012} & \textbf{0.7144} & \textcolor{green!50!black}{+0.3132} \\
\bottomrule
\end{tabular}%
}
\end{table*}

\subsection{Construction of UltraSAM3}
Based on the SAM3 architecture, we construct UltraSAM3 through full-parameter fine-tuning on ultrasound segmentation data. Unlike approaches that freeze the image or text encoders and only update task-specific modules, we update all trainable components of SAM3, including the image encoder, text encoder, detector, tracker, and segmentation-related modules. This strategy allows the model to adapt both its visual representations and concept grounding capability to ultrasound imaging, which is characterized by speckle noise, low contrast, acoustic shadowing, ambiguous boundaries, and strong appearance variations across organs and scanners.

Each training sample is represented as an ultrasound image--mask--concept triplet $(I_i, M_i, q_i)$. The concept prompt $q_i$ is derived from the corresponding anatomical or lesion category and is manually normalized when necessary to reduce ambiguity. The model is optimized to predict the target mask conditioned on the ultrasound image and the textual concept. The training objective combines mask reconstruction and region-level matching losses following the SAM3 segmentation paradigm:
\begin{equation}
    \theta^{*} =
    \arg\min_{\theta}
    \sum_{i=1}^{N}
    \mathcal{L}_{\mathrm{seg}}
    \big(f_{\theta}(I_i, q_i), M_i\big),
\end{equation}
where $\mathcal{L}_{\mathrm{seg}}$ denotes the segmentation loss used to supervise the predicted masks. Through full-parameter adaptation, UltraSAM3 learns ultrasound-aware visual features and improves the alignment between medical concepts and pixel-level segmentation targets.

\subsection{Instruction-guided Agent Framework}
Although UltraSAM3 can segment targets from textual concepts, directly feeding long and complex user instructions into the model may introduce redundant information or ambiguous expressions. To improve robustness under realistic user interaction, we introduce an instruction-guided agent as a front-end parser. Given an ultrasound image $I$, a complex user instruction $u$, the known organ domain $o$, and an organ-level candidate concept set $\mathcal{C}_{o} = \{c_1, c_2, \ldots, c_m\}$, the agent selects the most relevant target concept and rewrites the instruction into a concise concept prompt:
\begin{equation}
    q^{*} = g_{\phi}(I, u, o, \mathcal{C}_{o}),
\end{equation}
where $g_{\phi}$ denotes the agent and $q^{*}$ is the parsed SAM3-compatible prompt.

The parsed prompt $q^{*}$ is then used by UltraSAM3 for final mask prediction:
\begin{equation}
    \hat{M} = f_{\theta}(I, q^{*}).
\end{equation}
In practice, the agent outputs a structured response containing the selected organ, the chosen category identifier, the chosen category name, and the rewritten concept prompt. Importantly, the agent does not perform segmentation by itself; instead, it decomposes the task into high-level semantic instruction understanding and low-level mask prediction. This design preserves the segmentation ability of UltraSAM3 while improving its usability for complex natural-language queries.

\section{Experiment}

\subsection{Datasets and Evaluation Metrics}

For model training, we use 37 internal ultrasound segmentation datasets\cite{mclaughlan2024lung,marzola2021deep,belasso2020luminous,cunningham2020fallmud,orlando_abdomenus_ussimandsegm,zhao2022mmotu,ouyang2020video,leclerc2019deep,unityimaging2024echocardiography,wang2021echocp,reddy2023video,yang2023graphecho,shao2024micro,baum2023mr,stanfordaimi2024thyroidultrasoundcineclip,gong2023thyroid,gong2021multi,kronke2022tracked,pedraza2015open,xu_yiming_2022_7272660,ashkani2022fast,sappia2025acouslic,jieyun_2024_10829116,da2023fetal,songxiong2025focus,van2018automated,ding2022mallesnet,yap2017automated,al2020dataset,gomez2024bus,iqbal2024memory,ardakani2023open,guo2021segmentation,bi2023mi,jeevaws2025ultrasoundnormalkidney,singla2023open} collected from public sources. These datasets cover 13 anatomical categories, including lung, muscle, abdomen, ovary, heart, prostate, thyroid, liver, fetus, nerve, breast, carotid artery, and kidney. The training set contains 112,634 ultrasound images and 171,693 segmentation annotations. In addition, four external datasets are used for independent generalization evaluation, including BrEast~\cite{pawlowska2024curated}, CCAUI~\cite{momot2022commoncarotidartery}, 105US\_tumor~\cite{hann2017algorithm}, and KFGNet~\cite{wang2022key}. All experiments are conducted on a workstation equipped with four NVIDIA GeForce RTX 4090 GPUs.

We evaluate segmentation performance using Intersection over Union (IoU) and Dice coefficient, which measure the overlap between the predicted mask $\hat{M}$ and the ground-truth mask $M$. IoU is defined as
\begin{equation}
\mathrm{IoU} = \frac{|\hat{M} \cap M|}{|\hat{M} \cup M|},
\end{equation}
and Dice is defined as
\begin{equation}
\mathrm{Dice} = \frac{2|\hat{M} \cap M|}{|\hat{M}| + |M|}.
\end{equation}
Higher values of both metrics indicate better segmentation accuracy.




\subsection{Main Results on Representative Ultrasound Datasets}

To evaluate the effectiveness of UltraSAM3 across diverse ultrasound scenarios, we first conduct a representative benchmark on 13 anatomical categories. For each category, one representative dataset is selected from the training corpus, covering abdomen, breast, cardiac, carotid artery, fetal, kidney, liver, lung, muscle, nerve, ovarian, prostate, and thyroid ultrasound images. We compare UltraSAM3 with four concept- or text-driven biomedical segmentation baselines, including UniBiomed, BiomedParse, SAM3, and Medical SAM3. The results are reported using two widely used region-overlap metrics, IoU and Dice, which directly reflect pixel-level segmentation quality. As shown in Table~\ref{tab:representative_dataset_results}, UltraSAM3 achieves the best average performance, obtaining 0.6342 IoU and 0.7144 Dice.

Compared with the strongest baseline for each metric, UltraSAM3 brings absolute improvements of +0.3088 in IoU and +0.3132 in Dice on average. The gains are consistent across most anatomical categories, especially on challenging ultrasound targets such as thyroid, muscle, prostate, and cardiac structures. These results indicate that adapting SAM3 with ultrasound-specific image--mask--concept triplets substantially improves semantic grounding and segmentation accuracy under text-based concept prompts. The strong performance across heterogeneous organs and datasets further demonstrates the generalization ability of UltraSAM3 for universal ultrasound image segmentation.

\begin{figure}[!ht]
\centerline{\includegraphics[width=0.6\columnwidth]{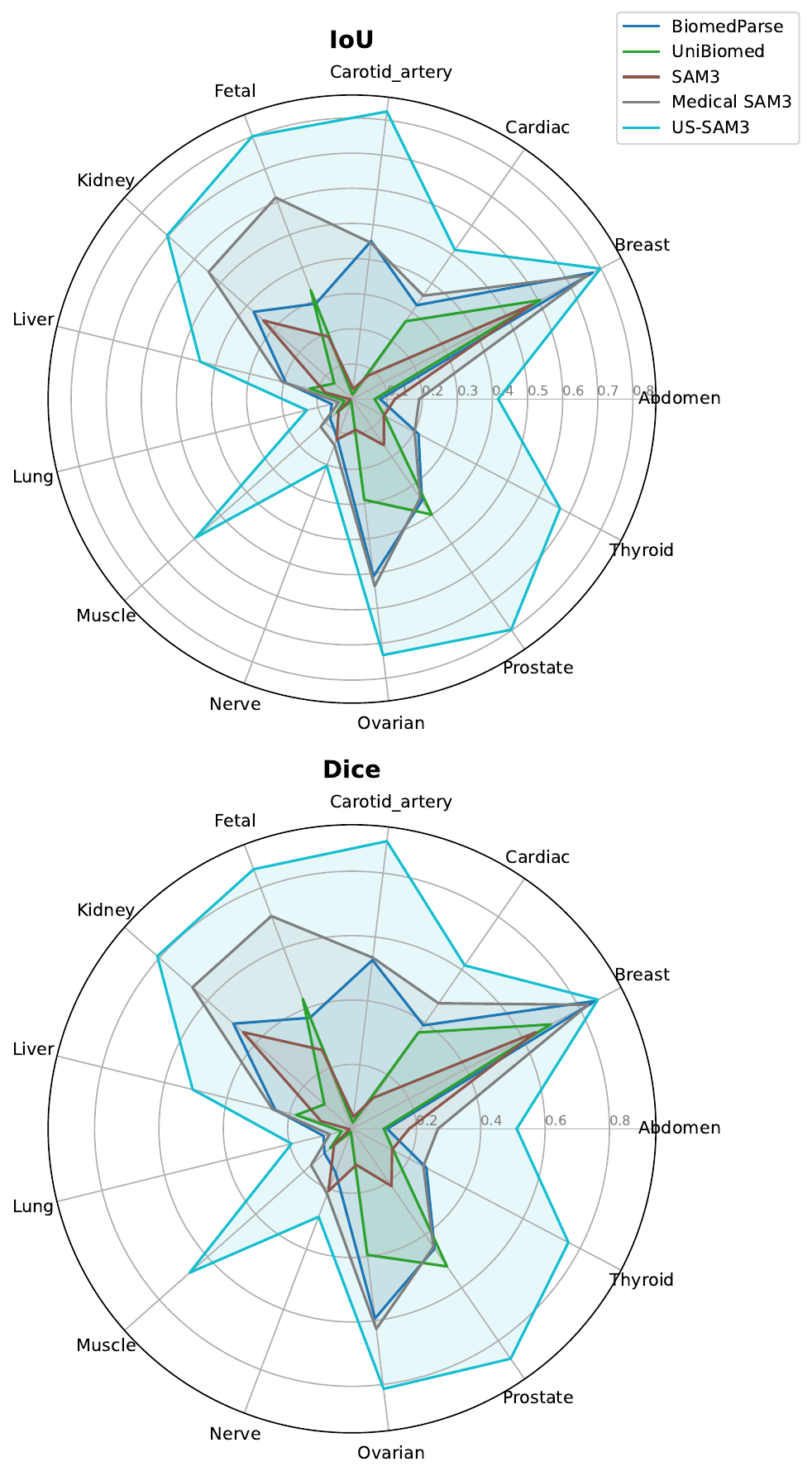}}
\caption{Organ-wise performance comparison on the full ultrasound corpus. All 37 datasets are grouped into 13 anatomical categories, and the results are averaged within each category when multiple datasets are available. Four radar plots report mAP, AP50, IoU, and Dice, respectively. UltraSAM3 shows consistently larger coverage across most organs and metrics, demonstrating its strong generalization ability for multi-organ ultrasound image segmentation.}
\label{fig:organ_wise_radar}
\end{figure}

\begin{figure}[!ht]
\centerline{\includegraphics[width=0.8\columnwidth]{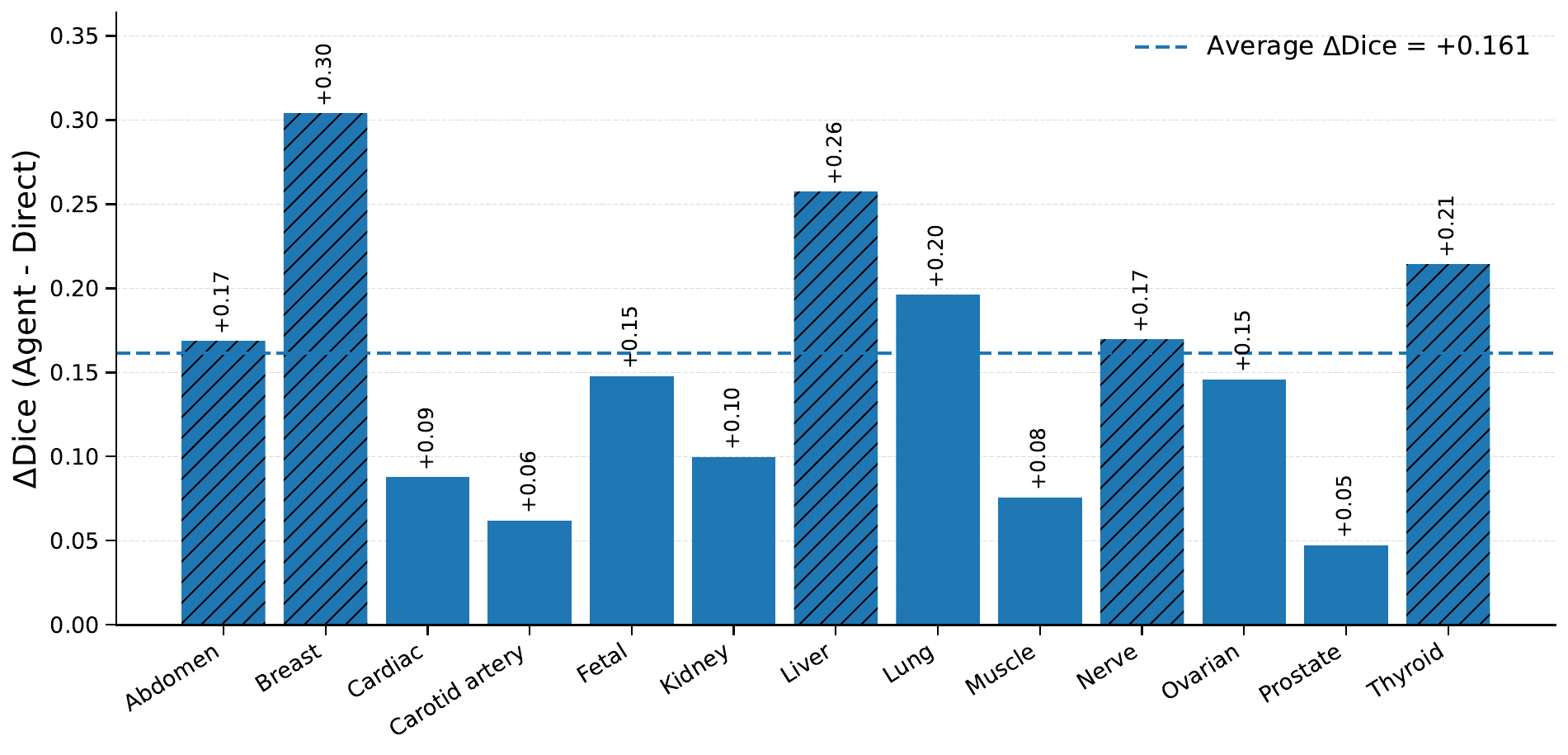}}
\caption{Dice improvement achieved by the instruction-guided agent over directly feeding complex instructions into UltraSAM3. For each dataset, 50 test images are randomly sampled, and complex user instructions are generated using Gemini 3.1 Pro. The bars represent the organ-wise average Dice gains obtained using agent-parsed concept prompts relative to direct complex-instruction inputs. Hatched bars indicate anatomical categories whose improvements are equal to or greater than the overall average. The dashed line denotes the average Dice improvement across all organs.}
\label{fig:agent_delta_dice}
\end{figure}

\subsection{Organ-wise Performance}

We further evaluate the robustness of UltraSAM3 on the full ultrasound corpus consisting of 37 datasets. Different from the representative-dataset experiment, this setting aggregates all datasets according to their anatomical categories. For categories containing multiple datasets, we report the average performance within the same organ group. This evaluation provides a more comprehensive view of model behavior across heterogeneous ultrasound sources, acquisition protocols, and annotation distributions.

As shown in Fig.~\ref{fig:organ_wise_radar}, UltraSAM3 consistently achieves the largest overall coverage in the radar plots across IoU and Dice. The advantage is especially clear for categories such as carotid artery, fetal, kidney, prostate, thyroid, and breast, where UltraSAM3 maintains strong region-overlap performance. Compared with general biomedical or generic concept-driven models, the consistent organ-wise gains suggest that ultrasound-specific adaptation is important for aligning noisy and low-contrast ultrasound appearances with medical concept prompts. These results further demonstrate that UltraSAM3 generalizes well across the complete multi-organ ultrasound corpus.

\begin{table*}[t]
\centering
\caption{Performance comparison on external ultrasound datasets. The best result is shown in bold, the second-best result is underlined, and $\Delta$ denotes the absolute change of UltraSAM3 compared with the strongest baseline.}
\label{tab:additional_dataset_results}
\setlength{\tabcolsep}{4pt}
\renewcommand{\arraystretch}{1.08}
\resizebox{\textwidth}{!}{%
\begin{tabular}{ll l ccccc c}
\toprule
\rowcolor{gray!15}
\textbf{Category} & \textbf{Dataset} & \textbf{Metric} & \textbf{UniBiomed} & \textbf{BiomedParse} & \textbf{SAM3} & \textbf{Medical SAM3} & \textbf{UltraSAM3} & \textbf{$\Delta$} \\
\midrule

\multirow{2}{*}{Breast} & \multirow{2}{*}{BrEast\cite{pawlowska2024curated}} 
& IoU  & 0.4118 & \underline{0.6378} & 0.4457 & 0.6245 & \textbf{0.6629} & \textcolor{green!50!black}{+0.0251} \\
& & Dice & 0.5204 & \underline{0.7429} & 0.5288 & 0.7066 & \textbf{0.7461} & \textcolor{green!50!black}{+0.0032} \\
\addlinespace[2pt]

\multirow{2}{*}{Carotid\_artery} & \multirow{2}{*}{CCAUI\cite{momot2022commoncarotidartery}} 
& IoU  & 0.0052 & \underline{0.6756} & 0.0257 & 0.5665 & \textbf{0.9133} & \textcolor{green!50!black}{+0.2377} \\
& & Dice & 0.0068 & \underline{0.7592} & 0.0403 & 0.6144 & \textbf{0.9543} & \textcolor{green!50!black}{+0.1951} \\
\addlinespace[2pt]

\multirow{2}{*}{Liver} & \multirow{2}{*}{105US\_tumor\cite{hann2017algorithm}} 
& IoU  & 0.0446 & 0.1552 & 0.0748 & \underline{0.2569} & \textbf{0.3264} & \textcolor{green!50!black}{+0.0695} \\
& & Dice & 0.0676 & 0.1890 & 0.0944 & \underline{0.2901} & \textbf{0.3642} & \textcolor{green!50!black}{+0.0741} \\
\addlinespace[2pt]

\multirow{2}{*}{Thyroid} & \multirow{2}{*}{KFGNet\cite{wang2022key}} 
& IoU  & 0.0474 & \underline{0.2002} & 0.0605 & 0.1279 & \textbf{0.4333} & \textcolor{green!50!black}{+0.2331} \\
& & Dice & 0.0628 & \underline{0.2331} & 0.0877 & 0.1577 & \textbf{0.5082} & \textcolor{green!50!black}{+0.2751} \\
\bottomrule
\end{tabular}%
}
\end{table*}

\subsection{Generalization on External Ultrasound Datasets}

To further assess the generalization ability of UltraSAM3, we evaluate it on several external ultrasound segmentation datasets that are not used for model training. These datasets cover different anatomical targets, including breast lesions, carotid artery, liver tumors, and thyroid nodules. Compared with the previous experiments on the main ultrasound corpus, this setting is more challenging because the model needs to handle unseen data sources, different acquisition conditions, and potential annotation distribution shifts. We compare UltraSAM3 with UniBiomed, BiomedParse, SAM3, and Medical SAM3 using mAP, AP50, IoU, and Dice.

As shown in Table~\ref{tab:additional_dataset_results}, UltraSAM3 achieves the best performance on all external datasets and all evaluation metrics. In particular, UltraSAM3 obtains consistent absolute improvements over the strongest baseline across breast, carotid artery, liver, and thyroid datasets. The gains are especially notable on CCAUI\_coco and KFGNet\_coco, indicating that the proposed ultrasound-specific concept adaptation can effectively transfer to unseen ultrasound domains. These results demonstrate that UltraSAM3 is not merely fitted to the training corpus, but learns robust ultrasound-aware semantic representations that support reliable concept-driven segmentation on external ultrasound images.

\begin{figure}[!ht]
    \centering
    \includegraphics[width=0.8\columnwidth]{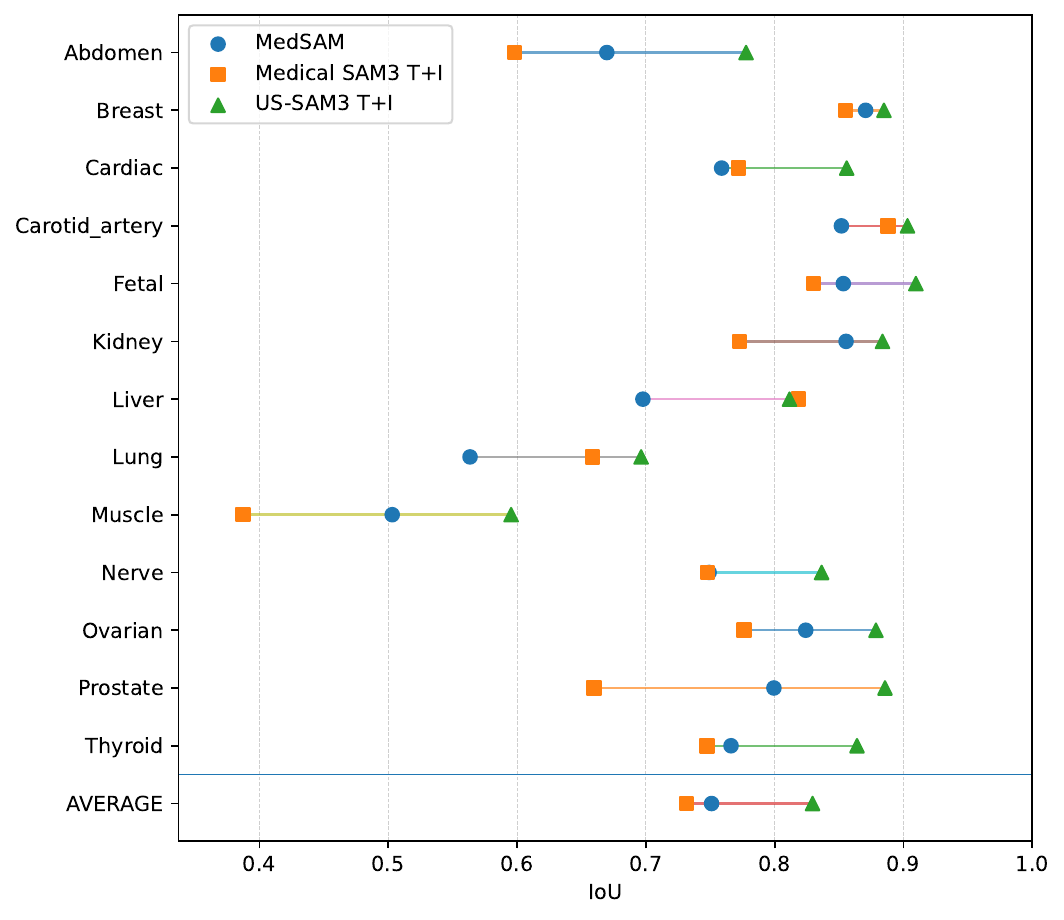}
    \caption{IoU comparison under visual-prompt-enhanced segmentation. Each row corresponds to one ultrasound anatomical category, and the connected markers indicate the IoU scores of MedSAM, Medical SAM3 with text and image prompts, and UltraSAM3 with text and image prompts. UltraSAM3 T+I achieves the highest average IoU and shows clear advantages across most organs, demonstrating that the proposed ultrasound-specific adaptation remains effective when visual prompts are available.}
    \label{fig:visual_prompt_iou}
\end{figure}

\begin{figure}[!ht]
\centerline{\includegraphics[width=\columnwidth]{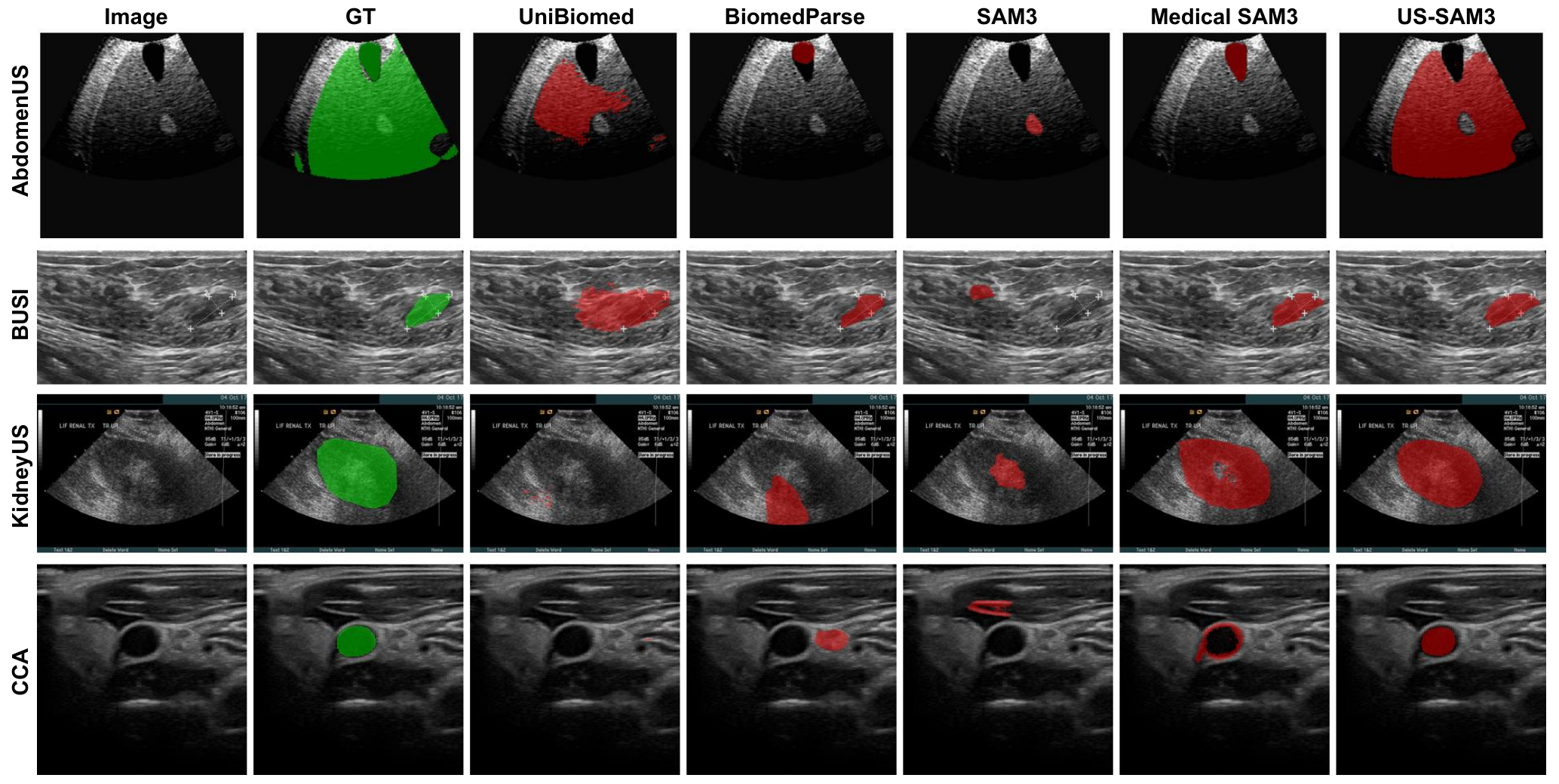}}
\caption{Qualitative comparison of different concept-driven segmentation models on representative ultrasound cases.}
\label{fig:case_study}
\end{figure}

\subsection{Effect of Visual Prompts}

Since SAM3-based models can incorporate both textual and visual prompts, we further evaluate whether the advantage of UltraSAM3 remains when visual guidance is available. In this experiment, we compare MedSAM, Medical SAM3 with text and image prompts, and UltraSAM3 with text and image prompts. To avoid introducing another large quantitative table, we report the organ-wise IoU results using a connected dot plot, which provides a compact visualization of the performance differences among visual-prompt-enhanced segmentation models.

As shown in Fig.~\ref{fig:visual_prompt_iou}, UltraSAM3 T+I achieves the highest average IoU and outperforms the competing methods on most anatomical categories. The improvement is particularly clear for abdomen, cardiac, carotid artery, fetal, kidney, nerve, ovarian, prostate, and thyroid categories. These results indicate that the ultrasound-specific adaptation of UltraSAM3 is not limited to text-only concept prompting. When additional visual prompts are provided, UltraSAM3 can still better align semantic concepts with ultrasound-specific visual patterns, leading to more accurate segmentation across diverse organs.

\subsection{Instruction-guided Agent for Complex User Queries}

Although UltraSAM3 supports concept-driven segmentation, directly using long or complex user instructions as text prompts may introduce ambiguity and weaken the localization ability of the model. To address this issue, we build an instruction-guided agent on top of UltraSAM3. For each dataset, we randomly sample 50 test images and use Gemini-3.1-Pro to generate complex but unambiguous user instructions according to the ground-truth target category. We then compare two inference strategies: directly feeding the complex instruction into UltraSAM3, and using the agent to parse the instruction into a concise target concept prompt before segmentation.

As shown in Fig.~\ref{fig:agent_delta_dice}, the agent consistently improves segmentation performance across all anatomical categories, achieving an average Dice gain of +0.161 over the direct complex-instruction baseline. The largest improvements are observed on breast, liver, thyroid, lung, and nerve ultrasound images, suggesting that complex natural-language instructions often contain redundant or indirect expressions that are not optimal for segmentation models. By explicitly identifying the target category and rewriting the query into a compact SAM3-compatible concept prompt, the agent improves the robustness of UltraSAM3 under realistic user interactions.

\subsection{Case Study}

Fig.~\ref{fig:case_study} presents representative qualitative results from different ultrasound categories. Compared with baselines, UltraSAM3 generates segmentation masks that are more spatially consistent with the ground-truth annotations. In abdominal and kidney ultrasound images, UltraSAM3 better captures the overall anatomical extent of the target region, while competing methods often produce incomplete or misplaced masks. In the breast and carotid artery cases, UltraSAM3 shows more accurate localization of small structures and lesion-like regions, demonstrating stronger robustness to low contrast, speckle noise, and ambiguous boundaries. These visual comparisons further support the quantitative results and indicate that ultrasound-specific concept adaptation improves both semantic grounding and pixel-level segmentation quality.

\section{Conclusion}

In this paper, we propose UltraSAM3, a concept-driven foundation model for universal ultrasound image segmentation. By adapting SAM3 with large-scale ultrasound image–mask–concept triplets, UltraSAM3 effectively improves semantic grounding and segmentation performance across diverse ultrasound organs and datasets. Experimental results demonstrate its strong accuracy, generalization ability, and robustness under both concept prompts and complex user instructions. In future work, we will consider designing more advanced agents that are better adapted to the model and ultrasound data, further improving the flexibility and clinical applicability of concept-driven ultrasound segmentation.

\section*{Acknowledgments}
This work is supported by the Science and Technology Project of Liaoning Province (2023JH2/101700363, 2024JH2/102600027), in part by the National Natural Science Foundation of China under Grant 62072073, 62106034, in part by the Fundamental Research Funds for the Central Universities under Grant DUT24ZD124, DUT25YG246, in part by the Dalian Innovation Fund 2021JJ12GX016, and the Science and Technology Project of Dalian City (2024JJ12GX025, 2023JJ12SN029 and 2023JJ11CG005).

\section*{REFERENCES}

\bibliographystyle{unsrt}  
\bibliography{references}  

\appendix

\section{Dataset Statistics}

Table~\ref{tab:appendix_dataset_statistics} provides detailed statistics of the 37 ultrasound datasets used for training US-SAM3. These datasets are grouped into 13 organ and anatomical categories, including abdomen, breast, cardiac, carotid artery, fetal, kidney, liver, lung, muscle, nerve, ovarian, prostate, and thyroid ultrasound. For each dataset, we report the number of training images, training masks, testing images, and testing masks. In total, the training corpus contains 112,634 ultrasound images and 171,693 segmentation masks, while the corresponding test splits contain 25,986 images and 40,939 masks. The dataset scale and anatomical diversity provide broad coverage for learning ultrasound-specific visual representations and concept-mask alignment across heterogeneous imaging scenarios.

\begin{figure}[!ht]
    \centering
    \includegraphics[width=0.6\columnwidth]{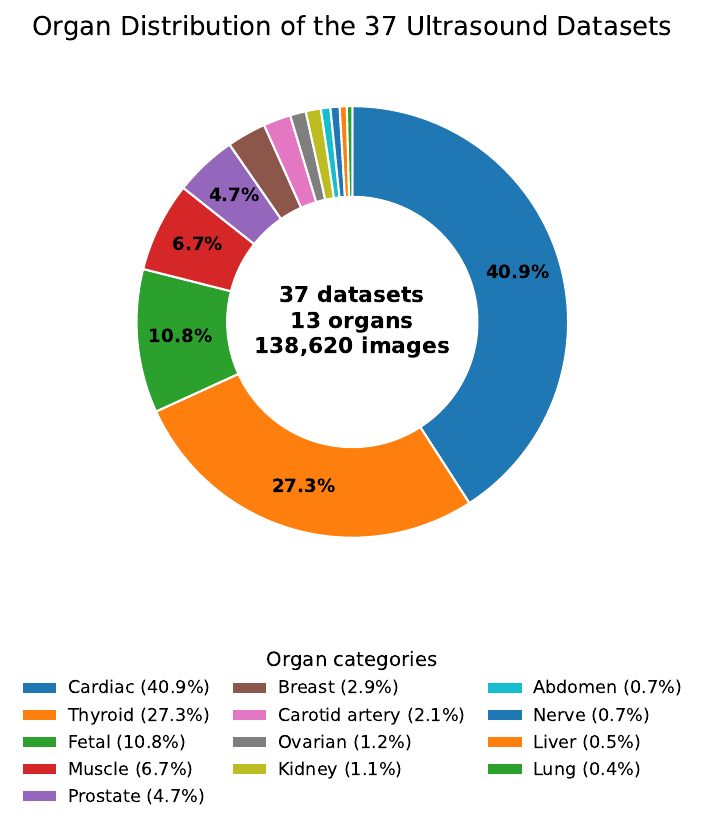}
    \caption{Organ-level data distribution of the 37 ultrasound datasets used in this work.}
    \label{fig:appendix_organ_distribution_pie}
\end{figure}

Figure~\ref{fig:appendix_organ_distribution_pie} provides an intuitive visualization of the organ-level composition of the training corpus. This figure aggregates the datasets by organ category and shows their relative proportions. The distribution is clearly imbalanced, with cardiac and thyroid ultrasound occupying the largest portions, followed by fetal, muscle, and prostate ultrasound. The remaining categories account for smaller proportions but increase the anatomical coverage of the corpus, which is important for evaluating the universality of concept-driven ultrasound segmentation.

\begin{table*}[t]
\centering
\caption{Statistics of the 37 ultrasound datasets used for training.}
\label{tab:appendix_dataset_statistics}
\setlength{\tabcolsep}{4pt}
\renewcommand{\arraystretch}{1.08}
\resizebox{\textwidth}{!}{%
\begin{tabular}{llrrrr}
\toprule
\rowcolor{gray!15}
\textbf{Organ} & \textbf{Dataset} & \textbf{Train Images} & \textbf{Train Masks} & \textbf{Test Images} & \textbf{Test Masks} \\
\midrule
Abdomen & AbdomenUS\cite{orlando_abdomenus_ussimandsegm} & 788 & 6,158 & 198 & 1,562 \\
Breast & BUID\cite{ardakani2023open} & 185 & 370 & 47 & 94 \\
Breast & BUSI\cite{al2020dataset} & 624 & 1,056 & 156 & 274 \\
Breast & BUS\_BRA\cite{gomez2024bus} & 1,500 & 3,000 & 375 & 750 \\
Breast & BUS\_DatasetB\cite{yap2017automated} & 130 & 260 & 33 & 66 \\
Breast & BUS\_UC\cite{iqbal2024memory} & 648 & 1,294 & 163 & 326 \\
Breast & S1\cite{guo2021segmentation} & 192 & 388 & 9 & 18 \\
\addlinespace[2pt]
Cardiac & CAMUS\cite{leclerc2019deep} & 1,600 & 4,800 & 400 & 1,200 \\
Cardiac & CardiacUDC\cite{yang2023graphecho} & 1,717 & 5,892 & 533 & 1,754 \\
Cardiac & EchoCP\cite{wang2021echocp} & 434 & 1,696 & 132 & 521 \\
Cardiac & EchoNet\_Dynamic\cite{ouyang2020video} & 17,496 & 17,496 & 2,552 & 2,552 \\
Cardiac & EchoNet\_Pediatric\cite{reddy2023video} & 12,071 & 12,071 & 3,300 & 3,300 \\
Cardiac & Unity\cite{unityimaging2024echocardiography} & 13,136 & 5,675 & 3,284 & 1,368 \\
\addlinespace[2pt]
Carotid artery & CCA\cite{bi2023mi} & 2,307 & 2,195 & 540 & 540 \\
\addlinespace[2pt]
Fetal & ACOUSLIC\cite{sappia2025acouslic} & 5,327 & 5,327 & 1,293 & 1,293 \\
Fetal & FASS\cite{da2023fetal} & 1,270 & 10,012 & 318 & 2,506 \\
Fetal & Fast\_UNet\cite{ashkani2022fast} & 1,128 & 1,128 & 283 & 283 \\
Fetal & HC\cite{van2018automated} & 799 & 799 & 200 & 200 \\
Fetal & fh\_ps\cite{jieyun_2024_10829116} & 3,200 & 6,400 & 800 & 1,600 \\
Fetal & focus\cite{songxiong2025focus} & 250 & 500 & 50 & 100 \\
\addlinespace[2pt]
Kidney & KidneyUS\cite{singla2023open} & 388 & 388 & 98 & 98 \\
Kidney & Ultrasound\_Normal\_Kidney\cite{jeevaws2025ultrasoundnormalkidney} & 864 & 1,728 & 216 & 432 \\
\addlinespace[2pt]
Liver & Annotated\_Ultrasound\_Liver\cite{xu_yiming_2022_7272660} & 588 & 1,020 & 147 & 250 \\
Lung & LUSS\cite{mclaughlan2024lung} & 451 & 2,977 & 113 & 775 \\
\addlinespace[2pt]
Muscle & FALLMUD\cite{cunningham2020fallmud} & 650 & 655 & 163 & 326 \\
Muscle & LUMINOUS\cite{belasso2020luminous} & 272 & 306 & 69 & 80 \\
Muscle & STMUS\_NDA\cite{marzola2021deep} & 6,535 & 6,536 & 1,634 & 1,634 \\
\addlinespace[2pt]
Nerve & UPBD\cite{ding2022mallesnet} & 764 & 3,441 & 191 & 935 \\
\addlinespace[2pt]
Ovarian & OTU\_2d\cite{zhao2022mmotu} & 1,175 & 2,382 & 294 & 596 \\
Ovarian & OTU\_3d\cite{zhao2022mmotu} & 136 & 292 & 34 & 80 \\
\addlinespace[2pt]
Prostate & MicroSeg\cite{shao2024micro} & 2,079 & 2,086 & 542 & 542 \\
Prostate & RegPro\cite{baum2023mr} & 3,518 & 3,788 & 367 & 396 \\
\addlinespace[2pt]
Thyroid & DDTI\cite{pedraza2015open} & 509 & 1,020 & 128 & 256 \\
Thyroid & Segthy\cite{kronke2022tracked} & 10,675 & 19,187 & 2,062 & 3,482 \\
Thyroid & TG3K\cite{gong2023thyroid} & 2,868 & 6,112 & 717 & 1,544 \\
Thyroid & TN3K\cite{gong2021multi} & 2,794 & 6,126 & 699 & 1,514 \\
Thyroid & Thyroid\_US\_Cineclip\cite{stanfordaimi2024thyroidultrasoundcineclip} & 13,566 & 27,132 & 3,846 & 7,692 \\
\midrule
\textbf{Total} & -- & \textbf{112,634} & \textbf{171,693} & \textbf{25,986} & \textbf{40,939} \\
\bottomrule
\end{tabular}%
}
\end{table*}

\section{Details of the Instruction-guided Agent}

Although US-SAM3 supports concept-driven ultrasound segmentation with concise textual prompts, directly using long or complex user instructions may introduce redundant expressions and semantic ambiguity. For example, a user may ask ``Please segment the suspicious tumor region in this breast ultrasound image'' rather than providing a compact concept such as ``breast lesion''. To improve the robustness of US-SAM3 under realistic clinical interaction, we introduce a lightweight instruction-guided agent as a front-end query parser. The agent does not perform segmentation by itself. Instead, it rewrites a complex natural-language instruction into a concise SAM3-compatible ultrasound concept prompt, which is then used by US-SAM3 for mask prediction.

Given an ultrasound image $I$, a user instruction $u$, an organ or task domain $o$, and an organ-specific candidate concept set
\begin{equation}
    C_o = \{c_1,c_2,\ldots,c_m\},
\end{equation}
where $c_j$ denotes the $j$-th candidate ultrasound target concept and $m$ is the number of candidate concepts, the agent $g_{\phi}$ selects the most relevant concept and produces a simplified prompt:
\begin{equation}
    q^{\ast} = g_{\phi}(I,u,o,C_o),
\end{equation}
where $\phi$ denotes the agent parameters or its prompting rules, and $q^{\ast}$ is the parsed concept prompt used for US-SAM3 inference.

The candidate concept selection process can be written as:
\begin{equation}
    j^{\ast} = \arg\max_{1\leq j\leq m} A_{\phi}(I,u,o,c_j),
\end{equation}
where $A_{\phi}(\cdot)$ represents the agent scoring function that measures the semantic compatibility between the image, the user instruction, the organ domain, and each candidate concept. The selected candidate $c_{j^{\ast}}$ is then rewritten into a concise textual prompt:
\begin{equation}
    q^{\ast} = R_{\phi}(u,c_{j^{\ast}}),
\end{equation}
where $R_{\phi}(\cdot)$ denotes the instruction rewriting function. In practice, $q^{\ast}$ is encouraged to be a short noun phrase or a very short instruction, such as ``thyroid nodule'', ``breast lesion'', ``fetal head'', or ``common carotid artery intima-media region''.

The agent returns a structured JSON response:
\begin{equation}
    Y_{\phi} = \{id^{\ast}, n^{\ast}, q^{\ast}, r^{\ast}\},
\end{equation}
where $id^{\ast}$ is the selected option identifier, $n^{\ast}$ is the selected category name, $q^{\ast}$ is the final SAM3-compatible concept prompt, and $r^{\ast}$ is a short explanation of the parsing decision. The parsed prompt is then passed to US-SAM3:
\begin{equation}
    \{(\hat{M}_k,s_k)\}_{k=1}^{K}=f_{\theta}(I,q^{\ast}),
\end{equation}
where $f_{\theta}$ denotes US-SAM3, $\hat{M}_k$ is the $k$-th candidate mask, $s_k$ is its confidence score, and $K$ is the number of candidate masks. The final mask is selected according to the highest confidence score:
\begin{equation}
    \hat{M}=\hat{M}_{k^{\ast}}, \quad 
    k^{\ast}=\arg\max_{1\leq k\leq K}s_k.
\end{equation}

In addition, we optionally filter candidate masks according to their pixel areas. Let $|\hat{M}_k|$ denote the area of the $k$-th candidate mask, $\tau_{\min}$ denote the minimum area threshold, and $\tau_{\max}$ denote the maximum area ratio threshold. For an image with height $H$ and width $W$, the valid candidate set is defined as:
\begin{equation}
    \Omega=\{k \mid |\hat{M}_k|\geq \tau_{\min},\ 
    |\hat{M}_k|\leq \tau_{\max}HW\}.
\end{equation}
The final candidate can then be selected within $\Omega$:
\begin{equation}
    \hat{M}=\hat{M}_{k^{\ast}}, \quad 
    k^{\ast}=\arg\max_{k\in\Omega}s_k.
\end{equation}

\subsection{Agent Prompt Template}

The front-end agent is implemented as a multimodal instruction parser. It receives the input ultrasound image, the user's natural-language segmentation question, and an optional fixed list of valid ultrasound target categories. The system prompt used by the agent is shown below.

\begin{tcolorbox}[
    title={System prompt for the instruction-guided agent},
    label={box:agent_system_prompt},
    colback=gray!5,
    colframe=gray!60,
    fonttitle=\bfseries,
    breakable
]
\small
\begin{verbatim}
You are a lightweight front-end agent for 
SAM3 ultrasound segmentation.

You receive an image, a user's 
natural-language segmentation question,
and an optional fixed list of valid ultra-
sound target categories.

Your job is not to segment masks. Your job 
is to rewrite the user question
into one concise SAM3 text prompt that 
directly names the target anatomy/pathology. 

If the category list contains the target, 
choose exactly one option from it. 
If none clearly match, still produce the 
best short SAM3
prompt. 

Return ONLY valid JSON with keys: 
chosen_option_id, chosen_category_name, 
sam3_prompt, reason.
\end{verbatim}
\end{tcolorbox}

The user prompt template is shown below. Here, \texttt{question} denotes the original user instruction, and \texttt{target\_payload} denotes the candidate category list represented as a JSON array. Each candidate contains an option identifier, an original category name, and a normalized semantic meaning.

\begin{tcolorbox}[
    title={User prompt template for concept selection and prompt rewriting},
    label={box:agent_user_prompt},
    colback=gray!5,
    colframe=gray!60,
    fonttitle=\bfseries,
    breakable
]
\small
\begin{verbatim}
User segmentation question: {question}

Valid target categories as JSON array:
{target_payload}

Write sam3_prompt as a concise noun phrase 
or very short instruction under 14 words. 
Prefer concrete medical/ultrasound terms 
like 'thyroid nodule' instead of vague 
words like 'it'.
\end{verbatim}
\end{tcolorbox}

The expected output format is a JSON object. The field \texttt{sam3\_prompt} is used as the final textual concept input to US-SAM3.

\begin{tcolorbox}[
    title={Expected JSON output of the instruction-guided agent},
    label={box:agent_json},
    colback=gray!5,
    colframe=gray!60,
    fonttitle=\bfseries,
    breakable
]
\small
\begin{verbatim}
{
  "chosen_option_id": "3",
  "chosen_category_name": "breast lesion",
  "sam3_prompt": "breast lesion",
  "reason": "The user asks to segment a 
  suspicious tumor-like region in a breast 
  ultrasound image, which corresponds to 
  breast lesion."
}
\end{verbatim}
\end{tcolorbox}

\section{Full Evaluation on the Training Datasets}

\subsection{Additional Evaluation Metrics}

Besides IoU and Dice, we also report mAP and AP50 for instance-level segmentation evaluation. Given an IoU threshold $t$, the average precision is denoted as $\mathrm{AP}_{t}$, and AP50 is computed at $t=0.50$:
\[
\mathrm{AP50} = \mathrm{AP}_{0.50}.
\]
The mAP is the mean AP over multiple IoU thresholds from 0.50 to 0.95 with a step size of 0.05:
\[
\mathrm{mAP} = \frac{1}{10}\sum_{t \in \{0.50,0.55,\ldots,0.95\}} \mathrm{AP}_{t}.
\]

\subsection{Analysis of Full Dataset Results}

Tables~\ref{tab:full_training_dataset_results_part1}--\ref{tab:full_training_dataset_results_part3} report the full performance comparison on all 37 ultrasound datasets used for training. For each dataset, we compare UniBiomed, BiomedParse, SAM3, Medical SAM3, and US-SAM3 using four metrics: mAP, AP50, IoU, and Dice. The best result is highlighted in bold, the second-best result is underlined, and $\Delta$ denotes the absolute improvement of US-SAM3 over the strongest baseline.

Overall, US-SAM3 achieves the strongest performance across the majority of datasets and metrics. Averaged over the 37 datasets, US-SAM3 obtains 0.5799 mAP, 0.8089 AP50, 0.6572 IoU, and 0.7367 Dice, outperforming all compared baselines by a clear margin. In particular, compared with the strongest baseline for each dataset and metric, US-SAM3 achieves average improvements of +0.3120 in mAP, +0.3827 in AP50, +0.1883 in IoU, and +0.1826 in Dice. These results show that the proposed ultrasound-specific adaptation improves not only pixel-level overlap accuracy but also instance-level localization and mask quality.

The improvement is especially evident for mAP and AP50. US-SAM3 achieves the best mAP on all 37 datasets and the best AP50 on 35 out of 37 datasets, indicating that the model can reliably identify the target regions across diverse ultrasound scenarios. This is important for concept-driven segmentation, where the model must localize targets from textual medical concepts without relying on manually provided visual prompts. The consistent gains over SAM3 and Medical SAM3 further suggest that general concept-driven segmentation models are not sufficient for ultrasound images, while ultrasound-specific image--mask--concept adaptation substantially strengthens semantic grounding.

\begin{table*}[!t]
\centering
\caption{Full performance comparison on the 37 training datasets (part 1 of 3). The best result is shown in bold, the second-best result is underlined, and $\Delta$ denotes the absolute improvement of US-SAM3 over the strongest baseline.}
\label{tab:full_training_dataset_results_part1}
\setlength{\tabcolsep}{4pt}
\renewcommand{\arraystretch}{1.08}
\resizebox{\textwidth}{!}{%
\begin{tabular}{ll l ccccc c}
\toprule
\rowcolor{gray!15}
\textbf{Category} & \textbf{Dataset} & \textbf{Metric} & \textbf{UniBiomed} & \textbf{BiomedParse} & \textbf{SAM3} & \textbf{Medical SAM3} & \textbf{US-SAM3} & \textbf{$\Delta$} \\
\midrule
\multirow{4}{*}{Abdomen} & \multirow{4}{*}{AbdomenUS} & mAP & 0.0001 & 0.0249 & 0.0060 & \underline{0.0560} & \textbf{0.1700} & \textcolor{green!50!black}{+0.1140} \\
& & AP50 & 0.0006 & 0.0416 & 0.0110 & \underline{0.0760} & \textbf{0.2750} & \textcolor{green!50!black}{+0.1990} \\
& & IoU & 0.0646 & 0.0802 & 0.1224 & \underline{0.1916} & \textbf{0.4163} & \textcolor{green!50!black}{+0.2247} \\
& & Dice & 0.0989 & 0.1093 & 0.1779 & \underline{0.2672} & \textbf{0.5118} & \textcolor{green!50!black}{+0.2446} \\
\addlinespace[2pt]
\multirow{4}{*}{Breast} & \multirow{4}{*}{BUID} & mAP & 0.2924 & 0.5104 & 0.3300 & \underline{0.5580} & \textbf{0.6760} & \textcolor{green!50!black}{+0.1180} \\
& & AP50 & 0.6527 & 0.7205 & 0.4810 & \underline{0.7300} & \textbf{0.8650} & \textcolor{green!50!black}{+0.1350} \\
& & IoU & 0.6227 & \textbf{0.8162} & 0.5856 & 0.7890 & \underline{0.8119} & \textcolor{red!60!black}{-0.0043} \\
& & Dice & 0.7297 & \textbf{0.8924} & 0.6609 & 0.8728 & \underline{0.8809} & \textcolor{red!60!black}{-0.0115} \\
\addlinespace[2pt]
\multirow{4}{*}{Breast} & \multirow{4}{*}{BUSI} & mAP & 0.2110 & \underline{0.4532} & 0.1150 & 0.3870 & \textbf{0.5150} & \textcolor{green!50!black}{+0.0618} \\
& & AP50 & 0.4748 & \underline{0.6389} & 0.1620 & 0.5710 & \textbf{0.7160} & \textcolor{green!50!black}{+0.0771} \\
& & IoU & 0.5559 & \textbf{0.7950} & 0.3590 & 0.5426 & \underline{0.5912} & \textcolor{red!60!black}{-0.2038} \\
& & Dice & 0.6455 & \textbf{0.8665} & 0.4076 & 0.6092 & \underline{0.6570} & \textcolor{red!60!black}{-0.2095} \\
\addlinespace[2pt]
\multirow{4}{*}{Breast} & \multirow{4}{*}{BUS\_BRA} & mAP & 0.1295 & 0.3131 & 0.1620 & \underline{0.3710} & \textbf{0.6020} & \textcolor{green!50!black}{+0.2310} \\
& & AP50 & 0.3615 & 0.5783 & 0.2560 & \underline{0.5990} & \textbf{0.8230} & \textcolor{green!50!black}{+0.2240} \\
& & IoU & 0.4981 & 0.6989 & 0.4522 & \underline{0.7293} & \textbf{0.8156} & \textcolor{green!50!black}{+0.0863} \\
& & Dice & 0.6084 & 0.7975 & 0.5054 & \underline{0.8091} & \textbf{0.8838} & \textcolor{green!50!black}{+0.0747} \\
\addlinespace[2pt]
\multirow{4}{*}{Breast} & \multirow{4}{*}{BUS\_DatasetB} & mAP & 0.2281 & 0.4560 & 0.2240 & \underline{0.5220} & \textbf{0.6530} & \textcolor{green!50!black}{+0.1310} \\
& & AP50 & 0.4960 & 0.6963 & 0.3440 & \underline{0.7110} & \textbf{0.9180} & \textcolor{green!50!black}{+0.2070} \\
& & IoU & 0.5013 & 0.7551 & 0.6290 & \underline{0.8314} & \textbf{0.8504} & \textcolor{green!50!black}{+0.0190} \\
& & Dice & 0.5747 & 0.8382 & 0.6991 & \underline{0.8925} & \textbf{0.9163} & \textcolor{green!50!black}{+0.0238} \\
\addlinespace[2pt]
\multirow{4}{*}{Breast} & \multirow{4}{*}{BUS\_UC} & mAP & 0.2597 & 0.4033 & 0.3200 & \underline{0.5040} & \textbf{0.5450} & \textcolor{green!50!black}{+0.0410} \\
& & AP50 & 0.5890 & 0.5999 & 0.4670 & \textbf{0.7060} & \underline{0.7030} & \textcolor{red!60!black}{-0.0030} \\
& & IoU & 0.6117 & 0.7387 & 0.7270 & \underline{0.8215} & \textbf{0.8503} & \textcolor{green!50!black}{+0.0288} \\
& & Dice & 0.7123 & 0.8226 & 0.8022 & \underline{0.8907} & \textbf{0.9102} & \textcolor{green!50!black}{+0.0195} \\
\addlinespace[2pt]
\multirow{4}{*}{Breast} & \multirow{4}{*}{S1} & mAP & \underline{0.7153} & 0.5406 & 0.3010 & 0.5350 & \textbf{0.7760} & \textcolor{green!50!black}{+0.0607} \\
& & AP50 & \textbf{1.0000} & 0.7638 & 0.3950 & 0.6800 & \underline{1.0000} & \textcolor{green!50!black}{+0.0000} \\
& & IoU & \underline{0.8467} & 0.8418 & 0.7256 & 0.8320 & \textbf{0.8735} & \textcolor{green!50!black}{+0.0268} \\
& & Dice & \underline{0.9147} & 0.9103 & 0.7954 & 0.8994 & \textbf{0.9286} & \textcolor{green!50!black}{+0.0139} \\
\addlinespace[2pt]
\multirow{4}{*}{Cardiac} & \multirow{4}{*}{CAMUS} & mAP & 0.0029 & \underline{0.4642} & 0.0000 & 0.1250 & \textbf{0.7320} & \textcolor{green!50!black}{+0.2678} \\
& & AP50 & 0.0212 & \underline{0.6476} & 0.0000 & 0.3690 & \textbf{0.9930} & \textcolor{green!50!black}{+0.3454} \\
& & IoU & 0.1757 & \textbf{0.5638} & 0.0132 & 0.2964 & \underline{0.3353} & \textcolor{red!60!black}{-0.2285} \\
& & Dice & 0.2516 & \textbf{0.6145} & 0.0256 & 0.4408 & \underline{0.4989} & \textcolor{red!60!black}{-0.1156} \\
\addlinespace[2pt]
\multirow{4}{*}{Cardiac} & \multirow{4}{*}{CardiacUDC} & mAP & 0.0030 & 0.0218 & 0.0030 & \underline{0.0550} & \textbf{0.4060} & \textcolor{green!50!black}{+0.3510} \\
& & AP50 & 0.0176 & 0.0565 & 0.0080 & \underline{0.1380} & \textbf{0.6890} & \textcolor{green!50!black}{+0.5510} \\
& & IoU & 0.1147 & 0.1247 & 0.0506 & \textbf{0.2520} & \underline{0.2137} & \textcolor{red!60!black}{-0.0383} \\
& & Dice & 0.1657 & 0.1634 & 0.0786 & \textbf{0.3708} & \underline{0.3322} & \textcolor{red!60!black}{-0.0386} \\
\addlinespace[2pt]
\multirow{4}{*}{Cardiac} & \multirow{4}{*}{EchoCP} & mAP & 0.0000 & \underline{0.0647} & 0.0000 & 0.0210 & \textbf{0.4320} & \textcolor{green!50!black}{+0.3673} \\
& & AP50 & 0.0000 & \underline{0.1450} & 0.0020 & 0.0780 & \textbf{0.7590} & \textcolor{green!50!black}{+0.6140} \\
& & IoU & 0.0168 & 0.1272 & 0.0153 & \textbf{0.3561} & \underline{0.2791} & \textcolor{red!60!black}{-0.0770} \\
& & Dice & 0.0284 & 0.1514 & 0.0290 & \textbf{0.5104} & \underline{0.4312} & \textcolor{red!60!black}{-0.0792} \\
\addlinespace[2pt]
\multirow{4}{*}{Cardiac} & \multirow{4}{*}{EchoNet\_Dynamic} & mAP & 0.0386 & 0.0596 & 0.0070 & \underline{0.2640} & \textbf{0.7610} & \textcolor{green!50!black}{+0.4970} \\
& & AP50 & 0.2028 & 0.1975 & 0.0180 & \underline{0.6250} & \textbf{0.9900} & \textcolor{green!50!black}{+0.3650} \\
& & IoU & 0.4291 & 0.2579 & 0.1119 & \underline{0.5089} & \textbf{0.8463} & \textcolor{green!50!black}{+0.3374} \\
& & Dice & 0.5690 & 0.3376 & 0.1514 & \underline{0.6121} & \textbf{0.9145} & \textcolor{green!50!black}{+0.3024} \\
\addlinespace[2pt]
\multirow{4}{*}{Cardiac} & \multirow{4}{*}{EchoNet\_Pediatric} & mAP & 0.0578 & 0.0549 & 0.0170 & \underline{0.2210} & \textbf{0.6750} & \textcolor{green!50!black}{+0.4540} \\
& & AP50 & 0.2626 & 0.1760 & 0.0470 & \underline{0.5460} & \textbf{0.9880} & \textcolor{green!50!black}{+0.4420} \\
& & IoU & 0.4854 & 0.3010 & 0.2551 & \underline{0.5589} & \textbf{0.8307} & \textcolor{green!50!black}{+0.2718} \\
& & Dice & 0.6297 & 0.3868 & 0.3539 & \underline{0.6897} & \textbf{0.9042} & \textcolor{green!50!black}{+0.2145} \\

\bottomrule
\end{tabular}%
}
\end{table*}

\begin{table*}[!t]
\centering
\caption{Full performance comparison on the 37 training datasets (part 2 of 3). The best result is shown in bold, the second-best result is underlined, and $\Delta$ denotes the absolute improvement of US-SAM3 over the strongest baseline.}
\label{tab:full_training_dataset_results_part2}
\setlength{\tabcolsep}{4pt}
\renewcommand{\arraystretch}{1.08}
\resizebox{\textwidth}{!}{%
\begin{tabular}{ll l ccccc c}
\toprule
\rowcolor{gray!15}
\textbf{Category} & \textbf{Dataset} & \textbf{Metric} & \textbf{UniBiomed} & \textbf{BiomedParse} & \textbf{SAM3} & \textbf{Medical SAM3} & \textbf{US-SAM3} & \textbf{$\Delta$} \\
\midrule
\multirow{4}{*}{Cardiac} & \multirow{4}{*}{Unity} & mAP & 0.0189 & \underline{0.1680} & 0.0010 & 0.0670 & \textbf{0.5300} & \textcolor{green!50!black}{+0.3620} \\
& & AP50 & 0.1178 & \underline{0.4138} & 0.0040 & 0.1900 & \textbf{0.7480} & \textcolor{green!50!black}{+0.3342} \\
& & IoU & 0.3926 & \underline{0.5725} & 0.0399 & 0.1731 & \textbf{0.5901} & \textcolor{green!50!black}{+0.0176} \\
& & Dice & 0.5366 & \textbf{0.6890} & 0.0614 & 0.2177 & \underline{0.6186} & \textcolor{red!60!black}{-0.0704} \\
\addlinespace[2pt]
\multirow{4}{*}{Carotid artery} & \multirow{4}{*}{CCA} & mAP & 0.0002 & \underline{0.2539} & 0.0040 & 0.1330 & \textbf{0.6940} & \textcolor{green!50!black}{+0.4401} \\
& & AP50 & 0.0009 & \underline{0.5553} & 0.0120 & 0.4260 & \textbf{0.9800} & \textcolor{green!50!black}{+0.4247} \\
& & IoU & 0.0133 & \underline{0.4531} & 0.0291 & 0.4479 & \textbf{0.8245} & \textcolor{green!50!black}{+0.3714} \\
& & Dice & 0.0190 & 0.5280 & 0.0374 & \underline{0.5354} & \textbf{0.8998} & \textcolor{green!50!black}{+0.3644} \\
\addlinespace[2pt]
\multirow{4}{*}{Fetal} & \multirow{4}{*}{ACOUSLIC} & mAP & 0.0287 & 0.0131 & 0.0110 & \underline{0.2010} & \textbf{0.5430} & \textcolor{green!50!black}{+0.3420} \\
& & AP50 & 0.0994 & 0.0294 & 0.0230 & \underline{0.3120} & \textbf{0.6860} & \textcolor{green!50!black}{+0.3740} \\
& & IoU & \underline{0.3051} & 0.2102 & 0.2237 & 0.2530 & \textbf{0.8723} & \textcolor{green!50!black}{+0.5672} \\
& & Dice & \underline{0.4078} & 0.2851 & 0.3017 & 0.3415 & \textbf{0.9283} & \textcolor{green!50!black}{+0.5205} \\
\addlinespace[2pt]
\multirow{4}{*}{Fetal} & \multirow{4}{*}{FASS} & mAP & 0.0007 & 0.0044 & 0.0080 & \underline{0.0370} & \textbf{0.4380} & \textcolor{green!50!black}{+0.4010} \\
& & AP50 & 0.0042 & 0.0082 & 0.0220 & \underline{0.0820} & \textbf{0.8210} & \textcolor{green!50!black}{+0.7390} \\
& & IoU & 0.1612 & 0.1043 & 0.2396 & \underline{0.3015} & \textbf{0.3985} & \textcolor{green!50!black}{+0.0970} \\
& & Dice & 0.2363 & 0.1573 & 0.3543 & \underline{0.4514} & \textbf{0.5244} & \textcolor{green!50!black}{+0.0730} \\
\addlinespace[2pt]
\multirow{4}{*}{Fetal} & \multirow{4}{*}{Fast\_UNet} & mAP & 0.1678 & 0.1410 & 0.1280 & \underline{0.5240} & \textbf{0.8970} & \textcolor{green!50!black}{+0.3730} \\
& & AP50 & 0.4115 & 0.2664 & 0.1750 & \underline{0.6070} & \textbf{0.9600} & \textcolor{green!50!black}{+0.3530} \\
& & IoU & 0.5528 & 0.5375 & 0.2834 & \underline{0.8677} & \textbf{0.9405} & \textcolor{green!50!black}{+0.0728} \\
& & Dice & 0.6882 & 0.6440 & 0.3601 & \underline{0.9250} & \textbf{0.9691} & \textcolor{green!50!black}{+0.0441} \\
\addlinespace[2pt]
\multirow{4}{*}{Fetal} & \multirow{4}{*}{HC} & mAP & 0.0958 & 0.2406 & 0.0830 & \underline{0.8680} & \textbf{0.9500} & \textcolor{green!50!black}{+0.0820} \\
& & AP50 & 0.3723 & 0.4418 & 0.1330 & \underline{0.9980} & \textbf{1.0000} & \textcolor{green!50!black}{+0.0020} \\
& & IoU & 0.5165 & 0.5048 & 0.1996 & \underline{0.8673} & \textbf{0.9379} & \textcolor{green!50!black}{+0.0706} \\
& & Dice & 0.6595 & 0.6039 & 0.2683 & \underline{0.9243} & \textbf{0.9655} & \textcolor{green!50!black}{+0.0412} \\
\addlinespace[2pt]
\multirow{4}{*}{Fetal} & \multirow{4}{*}{fh\_ps} & mAP & 0.0000 & 0.0000 & 0.0010 & \underline{0.3900} & \textbf{0.7520} & \textcolor{green!50!black}{+0.3620} \\
& & AP50 & 0.0000 & 0.0000 & 0.0040 & \underline{0.4970} & \textbf{0.9950} & \textcolor{green!50!black}{+0.4980} \\
& & IoU & 0.0025 & 0.0000 & 0.0455 & \underline{0.7624} & \textbf{0.7831} & \textcolor{green!50!black}{+0.0207} \\
& & Dice & 0.0042 & 0.0000 & 0.0773 & \underline{0.8637} & \textbf{0.8680} & \textcolor{green!50!black}{+0.0043} \\
\addlinespace[2pt]
\multirow{4}{*}{Fetal} & \multirow{4}{*}{focus} & mAP & \underline{0.1137} & 0.0574 & 0.0290 & 0.1090 & \textbf{0.7360} & \textcolor{green!50!black}{+0.6223} \\
& & AP50 & \underline{0.2995} & 0.1707 & 0.0560 & 0.2050 & \textbf{0.9870} & \textcolor{green!50!black}{+0.6875} \\
& & IoU & 0.4538 & 0.3856 & 0.1473 & \underline{0.6292} & \textbf{0.8684} & \textcolor{green!50!black}{+0.2392} \\
& & Dice & 0.5936 & 0.5215 & 0.2090 & \underline{0.7337} & \textbf{0.9182} & \textcolor{green!50!black}{+0.1845} \\
\addlinespace[2pt]
\multirow{4}{*}{Kidney} & \multirow{4}{*}{KidneyUS} & mAP & 0.0001 & 0.2083 & 0.0180 & \underline{0.6850} & \textbf{0.7400} & \textcolor{green!50!black}{+0.0550} \\
& & AP50 & 0.0006 & 0.4392 & 0.0370 & \underline{0.9530} & \textbf{0.9600} & \textcolor{green!50!black}{+0.0070} \\
& & IoU & 0.0964 & 0.5407 & 0.1959 & \underline{0.7643} & \textbf{0.8246} & \textcolor{green!50!black}{+0.0603} \\
& & Dice & 0.1569 & 0.6561 & 0.2627 & \underline{0.8515} & \textbf{0.8898} & \textcolor{green!50!black}{+0.0383} \\
\addlinespace[2pt]
\multirow{4}{*}{Kidney} & \multirow{4}{*}{Ultrasound\_Normal\_Kidney} & mAP & 0.0000 & 0.1389 & 0.0570 & \underline{0.3730} & \textbf{0.7960} & \textcolor{green!50!black}{+0.4230} \\
& & AP50 & 0.0000 & 0.3322 & 0.2780 & \underline{0.6400} & \textbf{0.9990} & \textcolor{green!50!black}{+0.3590} \\
& & IoU & 0.0380 & 0.2082 & \underline{0.4787} & 0.3267 & \textbf{0.5814} & \textcolor{green!50!black}{+0.1027} \\
& & Dice & 0.0720 & 0.3287 & \underline{0.6425} & 0.4738 & \textbf{0.7271} & \textcolor{green!50!black}{+0.0846} \\
\addlinespace[2pt]
\multirow{4}{*}{Liver} & \multirow{4}{*}{Annotated\_Ultrasound\_Liver} & mAP & 0.0024 & 0.0225 & 0.0080 & \underline{0.0630} & \textbf{0.2690} & \textcolor{green!50!black}{+0.2060} \\
& & AP50 & 0.0119 & 0.0505 & 0.0250 & \underline{0.1150} & \textbf{0.4250} & \textcolor{green!50!black}{+0.3100} \\
& & IoU & 0.1243 & 0.1941 & 0.0790 & \underline{0.2050} & \textbf{0.4454} & \textcolor{green!50!black}{+0.2404} \\
& & Dice & 0.1785 & 0.2455 & 0.0992 & \underline{0.2546} & \textbf{0.5099} & \textcolor{green!50!black}{+0.2553} \\
\addlinespace[2pt]
\multirow{4}{*}{Lung} & \multirow{4}{*}{LUSS} & mAP & 0.0047 & 0.0033 & \underline{0.0060} & 0.0010 & \textbf{0.2670} & \textcolor{green!50!black}{+0.2610} \\
& & AP50 & 0.0079 & 0.0096 & \underline{0.0100} & 0.0030 & \textbf{0.5660} & \textcolor{green!50!black}{+0.5560} \\
& & IoU & 0.0219 & \underline{0.0595} & 0.0041 & 0.0386 & \textbf{0.1325} & \textcolor{green!50!black}{+0.0730} \\
& & Dice & 0.0332 & \underline{0.0914} & 0.0077 & 0.0713 & \textbf{0.1936} & \textcolor{green!50!black}{+0.1022} \\
\addlinespace[2pt]
\multirow{4}{*}{Muscle} & \multirow{4}{*}{FALLMUD} & mAP & 0.0000 & 0.0000 & 0.0140 & \underline{0.1030} & \textbf{0.3610} & \textcolor{green!50!black}{+0.2580} \\
& & AP50 & 0.0000 & 0.0000 & 0.0310 & \underline{0.2370} & \textbf{0.4920} & \textcolor{green!50!black}{+0.2550} \\
& & IoU & 0.0032 & 0.0043 & \underline{0.0602} & 0.0079 & \textbf{0.2337} & \textcolor{green!50!black}{+0.1735} \\
& & Dice & 0.0063 & 0.0082 & \underline{0.0998} & 0.0154 & \textbf{0.2991} & \textcolor{green!50!black}{+0.1993} \\
\bottomrule
\end{tabular}%
}
\end{table*}

\begin{table*}[! t]
\centering
\caption{Full performance comparison on the 37 training datasets (part 3 of 3). The best result is shown in bold, the second-best result is underlined, and $\Delta$ denotes the absolute improvement of US-SAM3 over the strongest baseline.}
\label{tab:full_training_dataset_results_part3}
\setlength{\tabcolsep}{4pt}
\renewcommand{\arraystretch}{1.08}
\resizebox{\textwidth}{!}{%
\begin{tabular}{ll l ccccc c}
\toprule
\rowcolor{gray!15}
\textbf{Category} & \textbf{Dataset} & \textbf{Metric} & \textbf{UniBiomed} & \textbf{BiomedParse} & \textbf{SAM3} & \textbf{Medical SAM3} & \textbf{US-SAM3} & \textbf{$\Delta$} \\
\midrule
\multirow{4}{*}{Muscle} & \multirow{4}{*}{LUMINOUS} & mAP & 0.0000 & 0.0000 & 0.0150 & \underline{0.2650} & \textbf{0.7150} & \textcolor{green!50!black}{+0.4500} \\
& & AP50 & 0.0000 & 0.0000 & 0.0400 & \underline{0.5460} & \textbf{0.9840} & \textcolor{green!50!black}{+0.4380} \\
& & IoU & 0.0285 & 0.0188 & 0.0900 & \underline{0.2632} & \textbf{0.7512} & \textcolor{green!50!black}{+0.4880} \\
& & Dice & 0.0515 & 0.0330 & 0.1207 & \underline{0.3575} & \textbf{0.8420} & \textcolor{green!50!black}{+0.4845} \\
\addlinespace[2pt]
\multirow{4}{*}{Muscle} & \multirow{4}{*}{STMUS\_NDA} & mAP & 0.0049 & 0.0035 & 0.0000 & \underline{0.0090} & \textbf{0.5080} & \textcolor{green!50!black}{+0.4990} \\
& & AP50 & 0.0190 & 0.0125 & 0.0000 & \underline{0.0370} & \textbf{0.7740} & \textcolor{green!50!black}{+0.7370} \\
& & IoU & 0.1465 & \underline{0.2264} & 0.0012 & 0.0871 & \textbf{0.8023} & \textcolor{green!50!black}{+0.5759} \\
& & Dice & 0.2142 & \underline{0.3019} & 0.0017 & 0.1371 & \textbf{0.8818} & \textcolor{green!50!black}{+0.5799} \\
\addlinespace[2pt]
\multirow{4}{*}{Nerve} & \multirow{4}{*}{UPBD} & mAP & 0.0000 & 0.0321 & 0.0030 & \underline{0.0350} & \textbf{0.3830} & \textcolor{green!50!black}{+0.3480} \\
& & AP50 & 0.0000 & 0.0526 & 0.0070 & \underline{0.0680} & \textbf{0.6380} & \textcolor{green!50!black}{+0.5700} \\
& & IoU & 0.0067 & 0.1190 & 0.1239 & \underline{0.1402} & \textbf{0.2031} & \textcolor{green!50!black}{+0.0629} \\
& & Dice & 0.0109 & 0.1442 & 0.2070 & \underline{0.2174} & \textbf{0.2923} & \textcolor{green!50!black}{+0.0749} \\
\addlinespace[2pt]
\multirow{4}{*}{Ovarian} & \multirow{4}{*}{OTU\_2d} & mAP & 0.0186 & \underline{0.3011} & 0.0170 & 0.2600 & \textbf{0.6280} & \textcolor{green!50!black}{+0.3269} \\
& & AP50 & 0.0650 & \underline{0.5058} & 0.0300 & 0.4150 & \textbf{0.8940} & \textcolor{green!50!black}{+0.3882} \\
& & IoU & 0.3199 & \underline{0.5289} & 0.1099 & 0.4843 & \textbf{0.7356} & \textcolor{green!50!black}{+0.2067} \\
& & Dice & 0.4307 & \underline{0.6157} & 0.1414 & 0.5717 & \textbf{0.8137} & \textcolor{green!50!black}{+0.1980} \\
\addlinespace[2pt]
\multirow{4}{*}{Ovarian} & \multirow{4}{*}{OTU\_3d} & mAP & 0.0296 & 0.2423 & 0.0410 & \underline{0.3410} & \textbf{0.4900} & \textcolor{green!50!black}{+0.1490} \\
& & AP50 & 0.0713 & 0.4666 & 0.0600 & \underline{0.5930} & \textbf{0.7900} & \textcolor{green!50!black}{+0.1970} \\
& & IoU & 0.2578 & 0.4880 & 0.0668 & \underline{0.5885} & \textbf{0.7329} & \textcolor{green!50!black}{+0.1444} \\
& & Dice & 0.3582 & 0.5710 & 0.0841 & \underline{0.6808} & \textbf{0.8142} & \textcolor{green!50!black}{+0.1334} \\
\addlinespace[2pt]
\multirow{4}{*}{Prostate} & \multirow{4}{*}{MicroSeg} & mAP & 0.0809 & 0.0623 & 0.0030 & \underline{0.3270} & \textbf{0.7890} & \textcolor{green!50!black}{+0.4620} \\
& & AP50 & 0.2556 & 0.1233 & 0.0080 & \underline{0.5430} & \textbf{0.9810} & \textcolor{green!50!black}{+0.4380} \\
& & IoU & \underline{0.4698} & 0.3303 & 0.0111 & 0.3051 & \textbf{0.8732} & \textcolor{green!50!black}{+0.4034} \\
& & Dice & \underline{0.6002} & 0.4218 & 0.0193 & 0.3965 & \textbf{0.9222} & \textcolor{green!50!black}{+0.3220} \\
\addlinespace[2pt]
\multirow{4}{*}{Prostate} & \multirow{4}{*}{RegPro} & mAP & 0.0297 & 0.0919 & 0.1110 & \underline{0.1450} & \textbf{0.5540} & \textcolor{green!50!black}{+0.4090} \\
& & AP50 & 0.1062 & 0.2612 & 0.3020 & \underline{0.3740} & \textbf{0.8820} & \textcolor{green!50!black}{+0.5080} \\
& & IoU & 0.3294 & 0.3717 & 0.3079 & \underline{0.3785} & \textbf{0.7224} & \textcolor{green!50!black}{+0.3439} \\
& & Dice & 0.4390 & 0.4815 & 0.4137 & \underline{0.4946} & \textbf{0.8135} & \textcolor{green!50!black}{+0.3189} \\
\addlinespace[2pt]
\multirow{4}{*}{Thyroid} & \multirow{4}{*}{DDTI} & mAP & 0.0040 & 0.0216 & 0.0050 & \underline{0.0850} & \textbf{0.3270} & \textcolor{green!50!black}{+0.2420} \\
& & AP50 & 0.0153 & 0.0584 & 0.0130 & \underline{0.1670} & \textbf{0.5360} & \textcolor{green!50!black}{+0.3690} \\
& & IoU & 0.1179 & 0.2403 & 0.1688 & \underline{0.2479} & \textbf{0.6908} & \textcolor{green!50!black}{+0.4429} \\
& & Dice & 0.1624 & 0.3039 & 0.2366 & \underline{0.3172} & \textbf{0.7877} & \textcolor{green!50!black}{+0.4705} \\
\addlinespace[2pt]
\multirow{4}{*}{Thyroid} & \multirow{4}{*}{Segthy} & mAP & 0.0008 & \underline{0.0300} & 0.0000 & 0.0080 & \textbf{0.6870} & \textcolor{green!50!black}{+0.6570} \\
& & AP50 & 0.0033 & \underline{0.0662} & 0.0010 & 0.0160 & \textbf{0.9060} & \textcolor{green!50!black}{+0.8398} \\
& & IoU & 0.0454 & 0.0893 & 0.0038 & \underline{0.1483} & \textbf{0.5188} & \textcolor{green!50!black}{+0.3705} \\
& & Dice & 0.0673 & 0.1075 & 0.0054 & \underline{0.1950} & \textbf{0.6117} & \textcolor{green!50!black}{+0.4167} \\
\addlinespace[2pt]
\multirow{4}{*}{Thyroid} & \multirow{4}{*}{TG3K} & mAP & 0.0004 & 0.0000 & \underline{0.0020} & 0.0020 & \textbf{0.6350} & \textcolor{green!50!black}{+0.6330} \\
& & AP50 & 0.0015 & 0.0000 & \underline{0.0060} & 0.0060 & \textbf{0.9110} & \textcolor{green!50!black}{+0.9050} \\
& & IoU & \underline{0.0273} & 0.0066 & 0.0131 & 0.0034 & \textbf{0.7857} & \textcolor{green!50!black}{+0.7584} \\
& & Dice & \underline{0.0390} & 0.0109 & 0.0236 & 0.0058 & \textbf{0.8687} & \textcolor{green!50!black}{+0.8297} \\
\addlinespace[2pt]
\multirow{4}{*}{Thyroid} & \multirow{4}{*}{TN3K} & mAP & 0.0146 & 0.1918 & 0.0440 & \underline{0.2880} & \textbf{0.5000} & \textcolor{green!50!black}{+0.2120} \\
& & AP50 & 0.0495 & 0.3724 & 0.0820 & \underline{0.4830} & \textbf{0.7710} & \textcolor{green!50!black}{+0.2880} \\
& & IoU & 0.2409 & \underline{0.4888} & 0.2405 & 0.4308 & \textbf{0.6891} & \textcolor{green!50!black}{+0.2003} \\
& & Dice & 0.3225 & \underline{0.5781} & 0.3034 & 0.5152 & \textbf{0.7772} & \textcolor{green!50!black}{+0.1991} \\
\addlinespace[2pt]
\multirow{4}{*}{Thyroid} & \multirow{4}{*}{Thyroid\_US\_Cineclip} & mAP & 0.0031 & 0.0208 & 0.0000 & \underline{0.0490} & \textbf{0.3230} & \textcolor{green!50!black}{+0.2740} \\
& & AP50 & 0.0096 & 0.0479 & 0.0010 & \underline{0.0980} & \textbf{0.5240} & \textcolor{green!50!black}{+0.4260} \\
& & IoU & 0.1025 & \underline{0.2436} & 0.0870 & 0.1776 & \textbf{0.6625} & \textcolor{green!50!black}{+0.4189} \\
& & Dice & 0.1356 & \underline{0.3052} & 0.1339 & 0.2189 & \textbf{0.7520} & \textcolor{green!50!black}{+0.4468} \\
\bottomrule
\end{tabular}%
}
\end{table*}

\end{document}